\documentclass[10pt,twocolumn,letterpaper]{article}

\usepackage[pagenumbers]{cvpr} % To force page numbers, e.g. for an arXiv version

\usepackage{lato}
\usepackage[dvipsnames, table]{xcolor}

\newcommand{\model}[1]{\texttt{#1}}
\newcommand{\prompt}[1]{\textit{``#1''}}
 \renewcommand{\paragraph}[1]{\vspace{1ex}\noindent\textbf{#1}}

\definecolor{turquoise}{RGB}{20, 213, 200}
\definecolor{lightgray}{RGB}{230, 230, 230}
\definecolor{shadecolor}{rgb}{0.4,0.4,0.4}

\newcommand{\ours}{VPD\xspace}

\usepackage{makecell}
\usepackage{times}
\usepackage{epsfig}
\usepackage{float}
\usepackage{graphicx}
\usepackage{amsmath}
\usepackage{amssymb}
\usepackage{bbm}
\usepackage{booktabs}
\usepackage{bm}
\usepackage{multirow}
\usepackage{microtype}
\usepackage[T1]{fontenc}
\usepackage{soul}
\usepackage{subcaption} % Required for subfigures
\usepackage{listings}
\usepackage[toc,page]{appendix}
\usepackage[normalem]{ulem} % For uline

\usepackage{pifont}% http://ctan.org/pkg/pifont
\newcommand{\cmark}{\ding{51}}%
\newcommand{\xmark}{\ding{55}}%

\usepackage{etoolbox}% edit \@maketitle
\makeatletter
\let\@titlehook=\relax
\apptocmd{\@maketitle}{\@titlehook}{}{}
\newcommand{\titlehook}[1]{\def\@titlehook{#1}}
\makeatother

\definecolor{codegreen}{rgb}{0,0.6,0}
\definecolor{codegray}{rgb}{0.5,0.5,0.5}
\definecolor{codepurple}{rgb}{0.58,0,0.82}
\definecolor{backcolour}{rgb}{0.95,0.95,0.92}

\lstdefinestyle{mystyle}{
    backgroundcolor=\color{backcolour},   
    commentstyle=\color{codegreen},
    keywordstyle=\color{magenta},
    numberstyle=\tiny\color{codegray},
    stringstyle=\color{codepurple},
    basicstyle=\tiny,
    breakatwhitespace=false,         
    breaklines=true,                 
    captionpos=b,                    
    keepspaces=true,                 
    numbers=left,                    
    numbersep=5pt,                  
    showspaces=false,                
    showstringspaces=false,
    showtabs=false,                  
    tabsize=2
}
\definecolor{cvprblue}{rgb}{0.21,0.49,0.74}
\usepackage[pagebackref,breaklinks,colorlinks,citecolor=cvprblue]{hyperref}

\def\paperID{13148} % *** Enter the Paper ID here
\def\confName{CVPR}
\def\confYear{2024}

\title{Visual Program Distillation:\\ Distilling Tools and Programmatic Reasoning into Vision-Language Models}

\author{
     \textbf{Yushi Hu}$^{1, 2}\thanks{Work done while interning at Google Research.}$ \quad
     \textbf{Otilia Stretcu}$^{1}$\quad
     \textbf{Chun-Ta Lu}$^{1}$\quad
     \textbf{Krishnamurthy Viswanathan}$^{1}$\quad \\
     \textbf{Kenji Hata}$^{1}$\quad
     \textbf{Enming Luo}$^{1}$\quad
     \textbf{Ranjay Krishna}$^{2}$\quad
     \textbf{Ariel Fuxman}$^{1}$\quad\\
     $^1$Google Research\quad
     $^2$University of Washington\\
    {\fontsize{9pt}{10pt}\selectfont \ttfamily \textcolor{pink}{\url{https://visual-program-distillation.github.io/}}}
}

\begin{document}

% \titlehook{
% \begin{center}
%     \centering
%     \captionsetup{type=figure}
%     \includegraphics[width=0.9\linewidth]{figures/vpt_teaser_draft.jpg}
%         \captionof{figure}{We introduce Visual Program Tuning (VPT), which leverages visual programs to generate faithful chain-of-thought training data for large multimodal models (LMM). Left: PaLI trained with VPT is able to reason like a visual program, solving complex visual queries in an interpretable and compositional manner. Right: PaLI-3 VPT (5B) and PaLI-X VPT (55B) outperforms state-of-the-art vision-language generalist models on a wide range of tasks.
%         \ranjay{Make this figure 1 column. Remove the example and just showcase the results graph. That's the main takeaway. That the results are very strong across the board. Let's change the colors of the figure so that we use different shades of blue for all other methods. Use purple or something pleasant for VPD to distinguish it from the others.}
%         }
% \end{center}%
% }

\maketitle

\begin{abstract}
Solving complex visual tasks such as ``Who invented the musical instrument on the right?'' involves a composition of skills: understanding space, recognizing instruments, and also retrieving prior knowledge. Recent work shows promise by decomposing such tasks using a large language model (LLM) into an executable program that invokes specialized vision models.
However, generated programs are error-prone: they omit necessary steps, include spurious ones, and are unable to recover when the specialized models give incorrect outputs.
Moreover, they require loading multiple models, incurring high latency and computation costs. 
We propose \textbf{Visual Program Distillation (VPD)}, an instruction tuning framework that produces a vision-language model (VLM) capable of solving complex visual tasks with a single forward pass.
% Given a training dataset of complex visual tasks, 
VPD distills the reasoning ability of LLMs by using them to sample multiple candidate programs, which are then executed and verified to identify a correct one. It translates each correct program into a language description of the reasoning steps, which are then distilled into a VLM.
Extensive experiments show that VPD improves the VLM's ability to count, understand spatial relations, and reason compositionally. 
Our VPD-trained PaLI-X outperforms all prior VLMs, achieving state-of-the-art performance across complex vision tasks, including MMBench, OK-VQA, A-OKVQA, TallyQA, POPE, and Hateful Memes.
An evaluation with human annotators also confirms that VPD improves model response factuality and consistency.
Finally, experiments on content moderation demonstrate that VPD is also helpful for adaptation to real-world applications with limited data.
% Our analysis showcases that VPD improves model's ability to count, understand spatial relations, and reason compositionally.
% VPD can also quickly tune models for real-world applications with limited training data; our experiments on content moderation show that VPD is more sample efficient than traditional supervised training.

\end{abstract}

% \begin{figure}[ht]
% \centering
%   \vspace{-1ex}
%   \includegraphics[width=\linewidth]{figures/radar-chart.pdf}
%   \caption{
% We introduce \textit{Visual Program Distillation (VPD)}, a training framework which leverages LLM-generated programs to synthesis multimodal chain-of-thought training data for Vision-Language Models (VLMs). Our generalist models trained with VPD, \model{PaLI-3-VPD} \model{(5B)} and \model{PaLI-X-VPD} \model{(55B)}, outperform prior VLMs on a broad range of tasks, while producing human-interpretable and faithful reasoning steps.
% % \ranjay{This figure is very important. Right now, you can't tell which one is ours and which one is other people. I recommend making it salient by making our colors pop more. At the very least, add (ours) to the legend.} \otilia{We tried the suggestion of using greys, but you can't distinguish the models anymore. I added "ours". Is this fine?}
% }
%   \vspace{-4mm}
%   \label{fig:radar}
% \end{figure}

\begin{figure}[ht]
\centering
  \vspace{-2.5ex}
  \includegraphics[width=\linewidth]{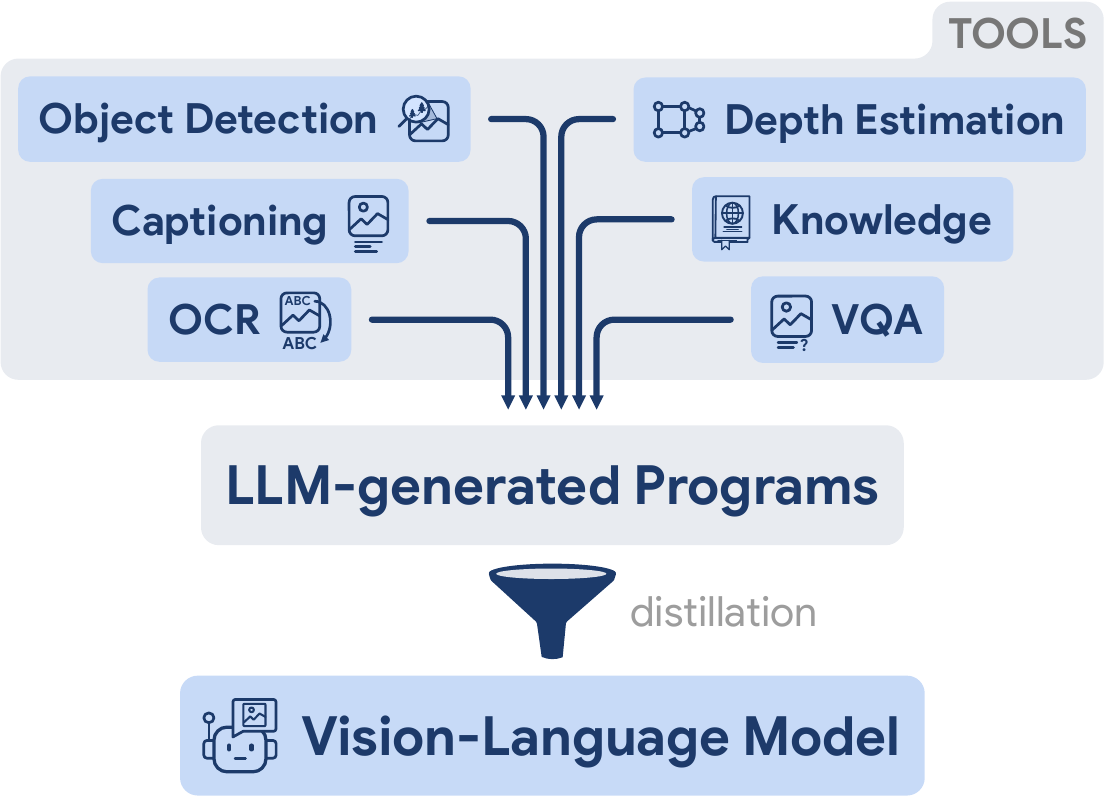}
   \vspace{-3ex}
  \caption{
We introduce \textit{Visual Program Distillation (VPD)}, a framework which leverages tools and LLM-generated programs to synthesize cross-modal reasoning data for vision-language model (VLM) training. 
VPD distills the skills in specialist models (\eg, detection, depth estimation) and LLM reasoning ability into VLMs.
Our generalist models trained with VPD, \model{PaLI-X-VPD} \model{(55B)}, outperform prior VLMs on a broad range of tasks, while producing human-interpretable and faithful reasoning steps.
}
  \vspace{-3ex}
  \label{fig:radar}
\end{figure}

\vspace{-1ex}
\section{Introduction}
\label{sec:intro}

Vision-language models (VLMs) have become the pretrained backbone for many computer vision tasks~\cite{alayrac2022flamingo, wang2022git, Li2023BLIP2BL, liu2023visual, zhu2023minigpt4, bai2023qwenvl,ye2023mplugowl2, wang2023cogvlm, chen2023shikra, Lu2022UnifiedIOAU, chen2022pali,chen2023palix,chen2023pali3}.
Yet, all these models still fall short of solving numerous visual reasoning tasks expected of competent vision models.
Even state-of-the-art (SOTA) proprietary vision-language models such as GPT-4V~\cite{openai2023gpt4} do not perform well on tasks that involve counting and spatial reasoning~\cite{yang2023dawn}.
They find it difficult to count (TallyQA~\cite{acharya2019tallyqa}), to compositionally reason (GQA~\cite{hudson2019gqa}), and to reason with external knowledge (OK-VQA~\cite{marino2019ok}, A-OKVQA~\cite{schwenk2022okvqa}).
Many of these tasks require VLMs to conduct compositional reasoning, which still remains an unsolved challenge.
For instance, answering the question \prompt{Who invented the musical instrument on the right?} involves a composition of skills: identifying objects, applying spatial reasoning to locate the one on the right, recognizing the musical instrument, and accessing prior knowledge to retrieve the inventor.
% Despite ongoing research, effectively incorporating compositional reasoning into vision-language models remains an unsolved challenge.

\begin{figure*}[t!]
\centering
  \includegraphics[width=\textwidth]{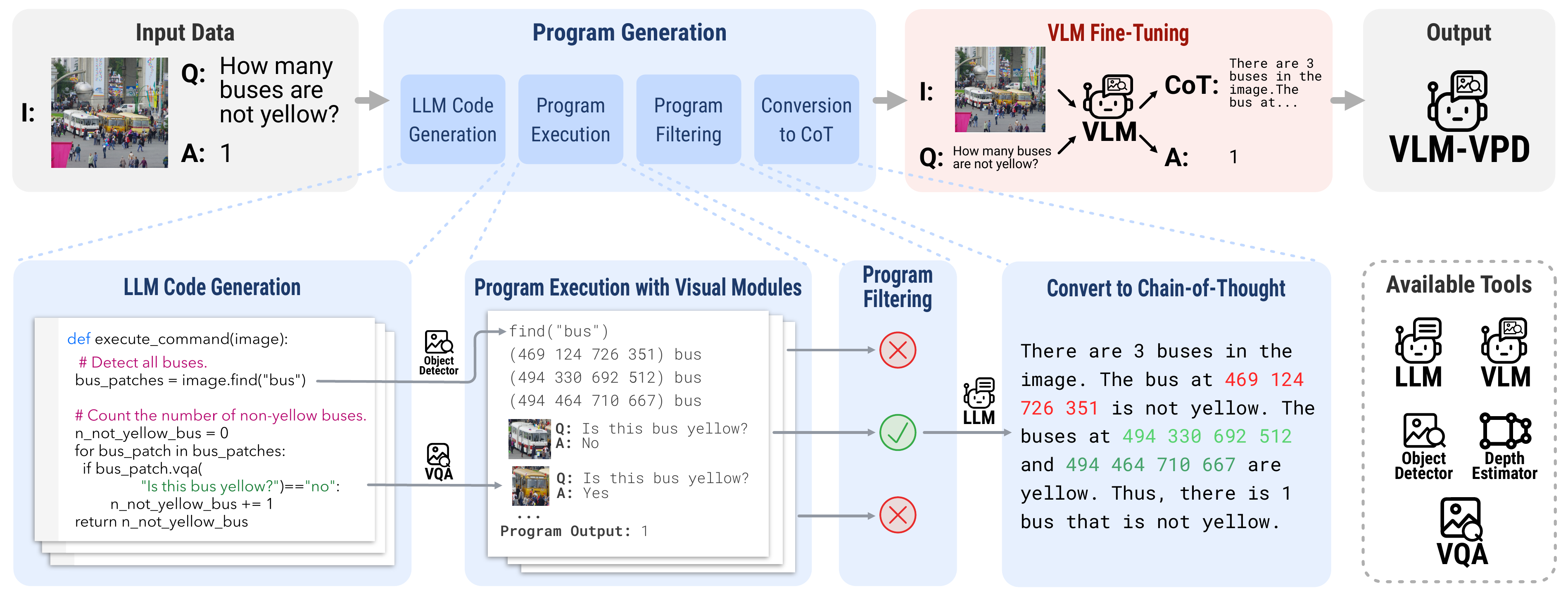}
  \vspace{-3ex}
  \caption{Overview of Visual Program Distillation (VPD). VPD uses an LLM and specialized vision tools to generate faithful chain-of-thought (CoT) training data for vision-language models (VLMs). Given a multimodal input, our 4-step data synthesis pipeline generates a CoT that answers the query.
  In the example above, our synthesized CoT contains a series of reasoning steps: find all buses, check if each bus is yellow, and aggregate the count into a final answer. The CoT also contains the grounding information given by object detection. 
  %By training VLMs with these CoTs, VPD distills the vision tools and  faithful programmatic reasoning into VLMs.
  }
  \vspace{-4mm}
  \label{fig:method}
\end{figure*}

In contrast, large language models (LLMs) have demonstrated remarkable performance at generating code that solves complex and compositional tasks~\cite{anil2023palm2, openai2023gpt4, chen2022program, xu2023lemur, lyu2023faithful}. 
% LLMs even produce reasoning steps to outline their ``chain-of-thought''~\cite{}.
Several recent papers~\cite{gupta2023visprog,surismenon2023vipergpt,hu2023avis,subramanian2023modular}
%, including last year's best paper at CVPR~\cite{gupta2023visprog}, 
capitalize on this by prompting LLMs to generate  programs where each step corresponds to a reasoning step.
The programs invoke specialized ``tools'' (or specialized vision models) to explicitly execute each reasoning step.
For the question above, the program would call an ``object detector'' tool to identify and isolate all the objects, a ``fine-grained object classification'' tool to recognize the musical instrument, and a ``knowledge-based question answering'' tool to retrieve its inventor.

Although innovative, generating explicit programs is computationally expensive in practice, prone to errors, and still underperforms end-to-end models.
Programs require loading and executing multiple tools, leading to high latency and computational cost.
Moreover, generated programs may omit necessary steps or include spurious ones. Even when the program is correct, vision model invocations can produce incorrect outputs, from which the overall program cannot recover.
Unfortunately, empirical results show that visual programs still fall short of end-to-end fine-tuned models~\cite{surismenon2023vipergpt}.

Another line of work is visual instruction tuning~\cite{liu2023visual}, which tries to distill the instruction-following ability of LLMs into VLMs.
They prompt LLMs with image captions and bounding box annotations in order to generate queries and answers that are used to fine-tune VLMs~\cite{liu2023visual,wang2023nonintrusive,chen2023shikra, hu2022promptcap}.
However, this approach has important limitations: image captions can miss fine-grained visual information, and LLM are prone to producing inconsistent outputs for custom vision representations like bounding boxes~\cite{Feng2023LayoutGPTCV}. As a result, existing instruction-tuned VLMs still struggle with tasks that requires complex visual reasoning~\cite{liu2023mmbench, liu2023improved, dai2023instructblip, chen2023shikra}.

In this work, we present \textbf{Visual Program Distillation (VPD)}, a novel distillation method that induces complex cross-modal reasoning capabilities into vision-language models (Fig.~\ref{fig:method}).
As the name suggests, VPD combines two key insights to deliver a training paradigm that surpasses the sum of its parts: It relies on (1) advancements in visual programs that use tools~\cite{gupta2023visprog} and (2) the recent breakthroughs in distillation through chain-of-thought reasoning~\cite{hsieh2023distilling}.
Given a labeled training dataset of complex visual tasks, VPD generates a correct program, and then distills its reasoning steps into vision-language models.
To avoid using programs that give the wrong answer, VPD prompts an LLM to generate multiple candidate programs, and executes every one of them. When labeled data is available, it then filters for programs that produce the correct answer upon execution. Therefore, our programs comprise multiple vision tools, are executable, and yield the correct answer when executed.
Next, VPD rewrites the correct programs as natural language chain-of-thought instructions and uses step-by-step distillation~\cite{hsieh2023distilling} to inject the reasoning abilities into VLMs.

Our best instruction-tuned model, \model{PaLI-X-VPD (55B)}, sets a new SOTA result on 8 classical VQA tasks and 2 zero-shot multimodal benchmarks. Our models even outperform the recent SOTA established by \model{PaLI-X}~\cite{chen2023palix}.
Importantly, we conduct a quality evaluation with human raters which shows that \model{PaLI-X-VPD} generates more consistent and faithful rationales compared to its counterpart trained using instruction-tuning data.
In addition, we experiment with \model{PaLI-3}\model{(5B)}, and show that VPD also improves the performance of smaller-scale models.
Finally, experiments on Hateful Memes~\cite{kiela2020hateful} show that VPD is also helpful for adapting to new tasks, even when no labels are available.
% We conclude by discussing potential ways for further extending VPD, including adding more powerful vision tools (e.g., segmentation), leveraging better code-generation models, and developing new filtering strategies that are applicable to large-scale unlabeled images. \otilia{Can we skip the last sentence? I think it's not needed for the intro.} \yushi{agree. Replace this with content moderation}

% \ranjay{I have said this before but I want to make sure I put this up here at the top. Every good paper makes 1 contribution. Not 2, not 3. Think of any top paper you remember. Can you remember all of their contributions? Or do you remember that 1 salient contribution? We need to stop using contribution lists in papers. It dilutes the paper.} 

% \ranjay{Here is another reason not to use lists. A contribution list with N items gives a reviewer N reasons to potentially reject a paper. A paper with a strong singular contribution makes it harder for a reviewer to reject.}

\section{Related work}
VPD is a general method for improving any vision-language model, and includes automatic program generation and training with chain-of-thought data as steps of the proposed framework. We discuss each of these research areas.

\paragraph{Vision-language models (VLMs).}
Most recent generative VLMs share a common structure: a pre-trained visual encoder, a pre-trained LLM, and a connector between the two modalities~\cite{alayrac2022flamingo, chen2022pali, wang2022git, Li2023BLIP2BL, liu2023visual, zhu2023minigpt4, bai2023qwenvl,ye2023mplugowl2, wang2023cogvlm, Lu2022UnifiedIOAU}.
The models are trained on large-scale image-text pairs from various tasks to adapt to both modalities.
They are also tuned on LLM-generated visual instructions to enable models to follow versatile instructions from users~\cite{liu2023visual}.
Some models also include bounding boxes in the pre-training data to improve the VLM on visually grounded tasks~\cite{zhang2022glipv2, peng2023kosmos2, wang2022ofa, chen2023shikra, Lu2022UnifiedIOAU, chen2023minigptv2, chen2023palix}. The bounding boxes are usually retrieved from COCO~\cite{lin2014microsoft} and Visual Genome~\cite{krishna2016visual}. Different from prior work, our method does not rely on provided dense annotations. We use LLM-generated code and specialized vision tools to generate our own visual instruction-tuning data.

\paragraph{Visual programming and agents.}
%Neuro-symbolic methods for vision tasks was pioneered by \citet{andreas2016neural}, who argue that complex vision tasks are compositional, and propose to solve them with perception units. 
With the advancement of large language models (LLMs)~\cite{chowdhery2022palm, anil2023palm2,brown2020language,openai2023gpt4,touvron2023llama}, recent work has started using LLMs as an interface to solving complex reasoning tasks with tools~\cite{zeng2022socratic, cheng2022binding, schick2023toolformer, shen2023hugginggpt, lu2023chameleon}, and also as an agent for vision tasks~\cite{yang2022empirical, hu2022promptcap, yang2023mmreact, hu2023avis}. From this line of work, the most relevant to us are VisProg~\cite{gupta2023visprog} and ViperGPT~\cite{surismenon2023vipergpt}, which leverage LLMs to generate executable programs with a sequence of invocations to specialized vision tools. They achieve SOTA zero-shot performance on various vision tasks, while being versatile and interpretable.

\paragraph{Training and inference with chain-of-thought.}
Chain-of-Thought (CoT)~\cite{wei2022chainofthought} has become a popular approach to improving LLM performance. Recent work such as Program-of-Thoughts (PoT)~\cite{chen2022program} and Faithful CoT~\cite{lyu2023faithful} further improve this framework by splitting inference into two steps: first generate a program with LLM, then execute the program. This approach achieves better accuracy and reduces hallucinations in the reasoning steps.
Moreover, CoT is also used to train language models. Distill step-by-step~\cite{hsieh2023distilling}, PaD~\cite{zhu2023pad}, and SCOTT~\cite{wang2023scott} train smaller language models with CoT generated by LLMs, showing that this can improve model performance and reasoning consistency. 
Mammoth~\cite{yue2023mammoth} trains an LLM with a hybrid of CoT and PoT rationales and achieves a SOTA model for math problems.
% Above works all serve as our inspirations.

\vspace{-0.6mm}
\section{Visual Program Distillation (VPD)}
\label{sec:method}
\vspace{-0.6mm}

We introduce \ours, a general model-agnostic framework that distills the reasoning power of LLM-generated programs together with the low-level image understanding abilities of vision tools into a single vision-language model (Fig.~\ref{fig:method}). At its core, VPD consists of two major steps:
\begin{enumerate}
    \item \textbf{Program generation and verification}: Given a textual query $q$ and a visual input $i$, VPD first generates a program that solves the query by making use of vision modules and tools, then converts the program execution trace into a chain-of-thought $c$ (\S\ref{sec:method:generate_cot}).
    \item \textbf{Distilling step-by-step}: The visual input $i$, textual query $q$, and chain-of-thought $c$ produced by the previous step are distilled into a vision-language model (VLM) (\S\ref{sec:method:multitask_tuning}).
\end{enumerate}

\subsection{Program generation and verification}
\label{sec:method:generate_cot}

Our training data synthesis pipeline is illustrated in the blue boxes of Fig.~\ref{fig:method}, and consists of four stages. 
% We follow the notation introduced by ViperGPT~\cite{surismenon2023vipergpt} to define them.
Given a sample $(i, q, y)$ consisting of a visual input $i$ and a textual query $q$ about its content and, when available, its ground truth answer $y$, we perform the following sequence of steps:
\begin{enumerate}
    \item {\em Program generation with LLM}: Given $q$, we generate a list of $k$ candidate programs $\pi(q) = \{z_1, z_2, ... , z_k\}$, where $\pi$ represents a program generation function. % (e.g., inference with an LLM).
    \item {\em Program execution with vision modules}: We execute each program $z_i$ with an execution engine $\phi$ to obtain its final result $\phi(i,z_i) = \hat y_i$. However, during program execution, we maintain the execution trace $t_i$ recording all the intermediate function calls and outputs. At the end of this step, we produce a list of programs, results, and execution traces $\{(z_1, \hat y_1, t_1),..., (z_k, \hat y_k, t_k) \}$.
    \item {\em Program filtering}: Among the $k$ candidate programs from the previous step, we keep a single tuple $(z, \hat y, t)$ with correct answer. %Details on filtering below.
    %We do so by first filtering out the programs whose answer $y_i$ does not match the ground truth human label included in the dataset. If many an \otilia{update descripton in this step if we include experiments on unlabeled data}
    \item {\em Converting program execution traces into chain-of-thought rationales}: We rewrite $t$ into a CoT $c$ using an LLM. %\ranjay{Can we move step 4 here to the next subsection. Let's have this subsection focus on just generating the correct program. The next subsection can convert the program and use it for distillation.}
\end{enumerate}

\noindent We now discuss in detail each of the steps above.

\paragraph{Program generation.}
% Recent work in LLMs has reached remarkable abilities at generating code~\cite{anil2023palm2, openai2023gpt4, xu2023lemur}. This capability is leveraged by methods such as VisProg~\cite{gupta2023visprog} and ViperGPT~\cite{surismenon2023vipergpt} to generate programs with vision modules as subroutines, that can tackle complex compositional visual tasks.
We adopt a similar approach with recent work~\cite{gupta2023visprog, surismenon2023vipergpt} in our program generation step, and use PaLM-2~\cite{anil2023palm2} as LLM to generate candidate programs for a given query $q$.
We prompt PaLM-2 with the same kind of text prompt as used by ViperGPT, which contains a detailed description of the available vision modules, followed by the query $q$ (prompt in Appendix~\S\ref{sec:appendix:prompts}).
The LLM is expected to directly output a Python function definition that will be executed in the following steps.
However, in our experiments, we find that the success rate of top-$k$ programs is much higher than the top-$1$ program.
Therefore, in contrast to prior work, which only samples one program $z$, when the ground truth answer $y$ is available, we set a temperature $T$ for LLM decoding and sample a list of top-$k$ candidate programs  $\{z_1, z_2, ... , z_k\}$ from the LLM. Then, we filter out one correct program $z$ in later steps. As shown in our ablation in \S\ref{sec:experiment:results}, this becomes crucial to our performance gain. We use $T=0.5$ and $k=5$ for all our experiments. For unlabeled data, $k=1$ and the filtering step can be skipped.

\paragraph{Program execution with vision modules.}
We use the same execution engine $\phi$ as ViperGPT~\cite{surismenon2023vipergpt}. An LLM-generated program $z_i$ is a Python function that takes the visual input $i$ as input. While the LLM outputs the program as a sequence of text, the execution engine $\phi$ is able to interpret this as a Python program and execute it, to obtain its return result $\hat y_i$. Additionally, $\phi$ also records the execution trace $t_i$ of the program $z_i$, which keeps a record of all the vision module calls, their inputs, and outputs. We use the following tools for the program: PaLI-X~\cite{chen2023palix} for simple visual queries, PaLI-X detection (distilled from OWLv2~\cite{minderer2023scaling}) for object detection; Google Cloud Depth API~\cite{Du2020DepthLab} for depth estimation, and PaLM-2~\cite{anil2023palm2} for external knowledge.

\paragraph{Program filtering.}
As we discussed in \S\ref{sec:intro}, visual programs are error-prone for a variety of reasons. 
% On one hand, the LLM-generated code might contain mistakes—which sometimes lead the program execution to fail (i.e., syntax mistakes), while in other cases contain reasoning mistakes that could pass undetected. On the other hand, even when the high level logic is correct, any of the modules or tools used as subroutines may return erroneous outputs (e.g., the object detection function fails to detect an object). Such errors are not recoverable during execution and are likely to lead to a wrong answer.
% With all these potential sources of error compounded, the overall performance of visual programs is much lower than that of supervised fine-tuned models.
The program might be wrong, and the execution process can introduce additional errors.
To overcome these issues, we employ a program filtering step. We start by sampling top-$k$ programs and attempt to execute each one.
During this process, any program that fails execution is instantly discarded.
We further filter the remaining programs, the strategy depending on the availability of labeled data.
For tasks where human labels are available, we select a single program per input sample whose answer is correct—that is, when the program output $\hat y_i$ matches the human label $y$. In this case our visual program pipeline acts like a CoT rationale annotator.
One potential problem is that some task answers are ambiguous.
For example, for a question like ``Where are the horses?'', the answers ``mountain'' and ``mountains'' are both correct. 
% To overcome such issues, 
We adopt the method in \cite{kamalloo2023evaluating} and use an LLM to determine if the program output is correct (details in \S\ref{sec:appendix:prompts}).
If there is more than one correct program per question, we select the top scoring one, according to the scores provided by the program-generating LLM $\phi$.
If no programs pass the test, the rated sample is not wasted—we simply use the correct answer $y$ as supervision in our fine-tuning stage, without an associated CoT.
When no human-rated answer is available, we directly use the top scoring executable program\footnote{Studying ways to approximate program correctness strategies for unlabeled data is an interesting direction for future work.}.

\paragraph{Converting program execution traces into chain-of-thought rationales.}
% \yushi{Add a plot of execution trace and COT examples}
After the filtering step, for each visual input $i$ and query $q$, we will have selected at most one program $z$ together with its execution result $\hat y$ and trace $t$. Since most existing VLMs have been pre-trained with text in natural language and not code, we use an LLM to rewrite the execution trace $t$ into a natural language CoT $c$ for our VLM distillation. Some examples are shown in \S\ref{sec:appendix:examples}. Concretely, similar to prior work~\cite{hu-etal-2022-context, hu2022promptcap}, we hand-craft $20$ examples of how an input ($q$, $z$, $t$) can be converted into a CoT, and use these as few-shots for prompting PaLM-2~\cite{anil2023palm2}, which performs in-context learning and generates CoTs for new examples. We include a concrete example in \S\ref{sec:appendix:prompts}.

% \begin{figure}[t!]
% \centering
%   \includegraphics[width=\linewidth]{figures/multitask-finetune.pdf}
%   \caption{
% Multitask distillation with synthesized COTs and program outputs.
% }
%   \vspace{-4mm}
%   \label{fig:multitask_training}
% \end{figure}

\subsection{Distilling step-by-step}
\label{sec:method:multitask_tuning}

% We first describe the existing framework for VLM training, and then extend it to incorporate rationales during training.
% We denote a dataset as $\mathcal{D} = \{I_i, x_i, y_i\}_{i=1}^N$, where each $I_i$ represent visual input, $x_i$ represent textual input, and $y_i$ represent the desired output label.
% Denote the VLM as f. We train $f$ to minimize the label prediction loss:
% \begin{equation}
%     \mathcal{L}_{label} = \frac{1}{N}\sum_{i=1}^N \ell(f(x_i), y_i)
% \end{equation}
% where $\ell$ is the sum of cross-entropy loss between the predict tokens and the target tokens. 

% \paragraph{Multi-task Learning with Rationales}

In this step, we fine-tune a backbone VLM with the training data generated in \S\ref{sec:method:generate_cot}, distilling the knowledge and reasoning steps of our generated programs into a single end-to-end model.
% We refer to ``fine-tuning'' in the broader sense—this can be done via parameter-efficient methods such as soft prompt-tuning~\cite{lester2021power} or LoRA~\cite{hu2021lora}, or end-to-end.
We do so in a multitask fashion, where the VLM is simultaneously fine-tuned on data synthesized for multiple types tasks (e.g., free-form VQA, multiple choice).

% \paragraph{Multitask training data.} 
Let $f$ represent our VLM model. While the same VLM architecture could solve all these tasks, it needs to be prompted differently to adapt to the task.
Following prior work~\cite{liu2023visual}, we manually design instructions for each task. For example, for free-form VQA queries, the instruction is \prompt{Answer with a single word or phrase}, while for multiple-choice queries we use \prompt{Answer with the option letter from the given choices directly}.
During fine-tuning, we combine the training samples generated for all tasks into a single dataset, so to account for the different types of tasks, we augment each sample with its corresponding task-specific prompt $p$.
% Let $T$ be the number of tasks, and we refer to the VLM prompted with the corresponding task instructions at $f_t$, for $t \in \{1,..,T\}$.

Using these instructions along with the data generated in \S\ref{sec:method:generate_cot}, we put together the training dataset  $\mathcal{D}=\{(i_j, q_j, \hat y_j, c_j, p_j)\}_{j=1}^N$, where
$N$ is the total number of samples,
$i_j$ is the visual input,
$q_j$ is the textual query,
$\hat y_i$ is the visual program output\footnote{When labeled data is available, this is equivalent to the ground truth label $y$ due to our filtering strategy.}, and
$c_j$ is the CoT rationale.

We train $f$ to minimize a loss for predicting both the label and the rationale.
As shown in the red box of Fig.~\ref{fig:method}, similar to~\cite{hsieh2023distilling}, we treat predicting the output label $\hat y_j$ and the rationale $c_j$ as two separate optimization goals.
However, since VLMs are open-ended text generation models, they need additional prompting to indicate whether we want a short answer or a long answer that includes a rationale. Therefore, we append the suffix $s_{c}=$\prompt{Explain the rationale to answer the question} at the end of the prompt for generating a CoT, and use the task instruction $p_j$ for short answers.
% We also denote such additional instructions for each sample as $r$. 
% We use $f_t^{CoT}$ to denote when the VLM is prompted to provide a rationale, and $f_t^{ans}$ when just the answer is required.
Our optimization objective is:
\begin{align}
\label{equation:loss}
    \mathcal{L} &= \mathcal{L}_{label} + \mathcal{L}_{rationale} \\
    &= \sum_{j=1}^N \ell(f(i_j, q_j, p_j), \hat y_j)+\ell(f(i_j, q_j, s_{c}), c_j)
\end{align}
Here $\ell$ is the cross entropy-loss normalized by sequence length to ensure that the labels (typically short) and rationales (typically long) have similar weights.
% $p$ is the task instruction for predicting label $y_i$. 
% Following prior work~\cite{liu2023visual}, the instruction is manually designed for each task. For example, for free-form VQA queries, the instruction is \textit{``Answer with a single word or phrase.''} For multiple-choice queries, the instruction is \textit{``Answer with the option letter from the given choices directly.''}
% $q$ is the task instruction for predicting the rationale $r_i$. It is \textit{``Explain the rationale to answer the question''} for all queries.
$\mathcal{L}_{rationale}$ both teaches the VLM to generate faithful reasoning steps similar to program execution traces, and carries more information beyond the label $\hat y$ that also helps the VLM in better predicting the label.
During test time, the rationale generation is not required.
We can adjust the task instruction to directly get the short output label using $p$, and using $s_{c}$ to get the human-interpretable reasoning steps if needed.

% \section{Experiments with PaLI-VPD: A Vision-Language Model Distilled with VPD}
\section{Experiments}
\label{sec:experiment}

\begin{figure}[t!]
\centering
  \vspace{-1em}
  \includegraphics[width=0.93\linewidth]{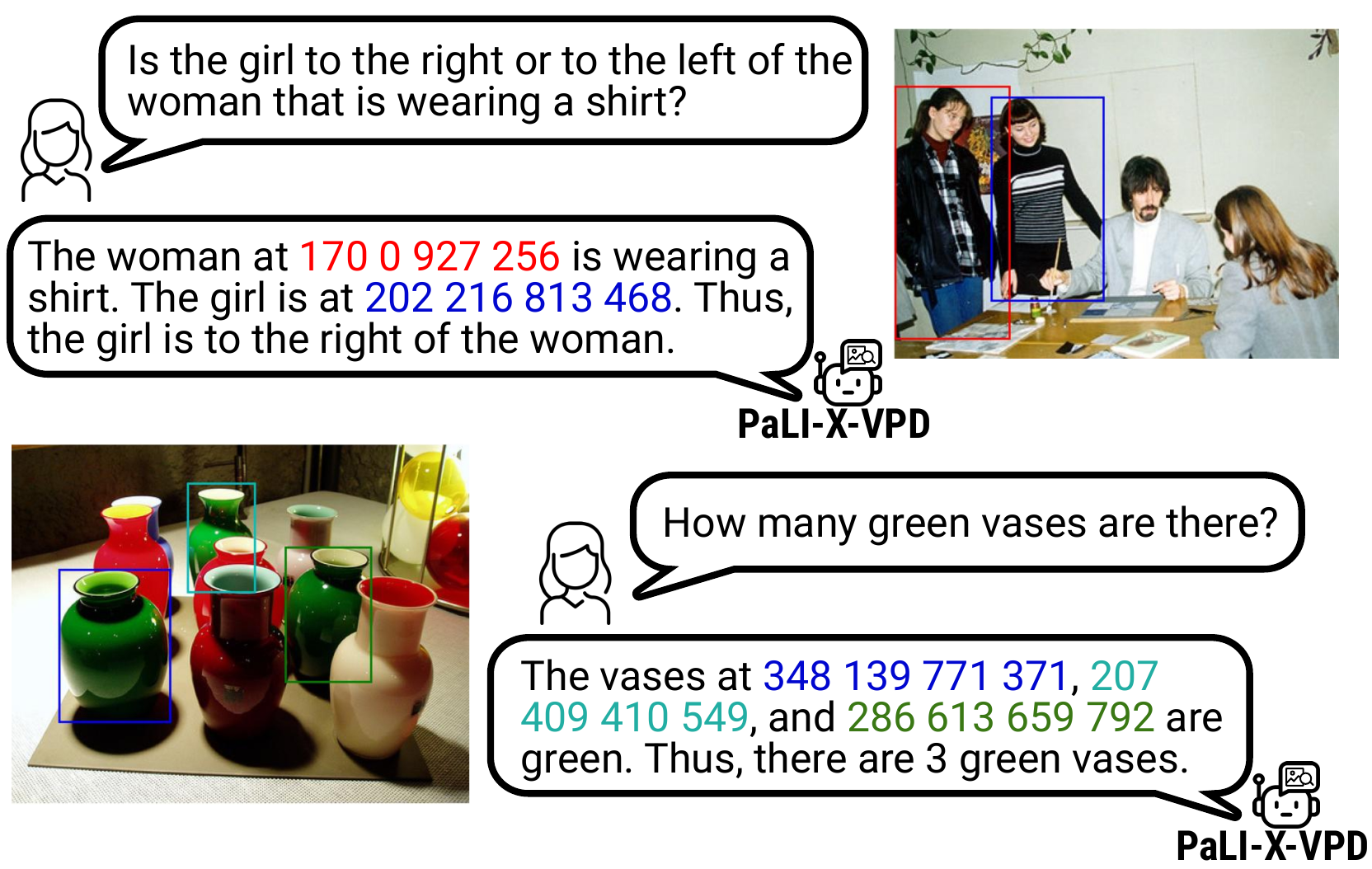}
   \vspace{-1ex}
  \caption{
Outputs of PaLI-X trained with VPD. Training with execution traces of filtered programs
% allows the model to produce human-interpretable and faithful reasoning steps. This ability also 
improves the model's ability to count, understand spatial relations, and reason compositionally. The images are from Visual Genome~\cite{krishna2016visual}.
}
  \vspace{-4mm}
  \label{fig:demo_intro}
\end{figure}

In this section, we demonstrate \ours's effectiveness by using it to train a generalist VLM.
We attempt this for two VLMs with different scales, \model{PaLI-3} \model{(5B)}~\cite{chen2023pali3} and \model{PaLI-X} \model{(55B)}~\cite{chen2023palix}. Detailed experimental setups are given in \S\ref{sec:experiment:setups}.
Qualitatively, models fine-tuned with \ours exhibit the ability to reason step-by-step like a program, as illustrated in Fig.~\ref{fig:demo_intro}, and supported by human evaluation results (\S\ref{sec:experiment:human_eval}).
Quantitative results show that our \model{PaLI-3-VPD} and \model{PaLI-X-VPD} achieve new SOTA on a broad range of tasks, on both generalist and per-task fine-tuning settings (\S\ref{sec:experiment:results}).
We also conduct a detailed analysis of the source of performance gain (\S\ref{sec:experiment:results}).
Finally, we conduct human evaluation on the quality of the rationales generated by our models, and compare it with rationales generated by an instruction-tuned model trained without \ours (\S\ref{sec:experiment:human_eval}).

\begin{table*}[h]
\small
\centering
\vspace{-1em}
{\fontsize{8.5pt}{9.5pt}\selectfont
\begin{tabular}{l|llllllll|cc}
\toprule[1.2pt]
% & \multicolumn{7}{|c|}{Instruction-tuning tasks} & \multicolumn{3}{c}{Zero-shot tasks} \\
% \midrule
Generalist Models & VQAv2 & GQA & OK-VQA & \multicolumn{2}{c}{A-OKVQA} & \multicolumn{2}{c}{TallyQA} & TextVQA & POPE & MMB \\
&  & & & MC & DA & Simp. & Comp. & & & \\
% \midrule
\hline
\rowcolor{lightgray}
\multicolumn{11}{l}{\textit{Prior generalist VLMs}} \\
\model{Flamingo} \model{(80B)}~\cite{alayrac2022flamingo} & 82.0 & - & 57.8* & - & - & - & -&  57.1 & - & - \\
% Unified-IO\textsubscript{XL} (3B) & 77.9* & - & 54.0* & - & - & - & - & - & - & -  \\
% KOSMOS-2 (1.6B) & 45.6 & - & - & - & - & - & - & - & - & 59.2 \\-
% \model{IDEFICS} \model{(80B)} & 60.0* & 45.2*  & - & - & - & - & - & 30.9* & - & 54.5 \\
% \model{BLIP-2 FlanT5\textsubscript{XXL}}\model{(12B)} & 65.0* & 32.3* & 45.9* & - & - & - &  - &  42.4* & - & - \\
\model{MiniGPT-4} \model{(Vicuna-13B)}~\cite{zhu2023minigpt4} & - & 43.5 & - & 67.2 & - & - & - & - &  - & 42.3 \\
\model{InstructBLIP} \model{(Vicuna-13B)}~\cite{dai2023instructblip} & - & 49.5* & - &-  & - & 75.2* & 57.5* & 50.7* & 78.9 &  44.0 \\
\model{Shikra} \model{(Vicuna-13B)}~\cite{chen2023shikra} &  77.4 & - & 47.2 & - & - & - & - & - & 84.7 & 58.8 \\
\model{Qwen-VL} \model{(9.7B)}~\cite{bai2023qwenvl} & 78.8 & 59.3 & 58.6 & - & - & 82.6* & 65.8* & 63.8 & - & 38.2 \\
\model{Qwen-VL-Chat} \model{(9.7B)}~\cite{bai2023qwenvl} & 78.2 & 57.5 & 56.6 & -  & - & 81.1* & 64.0* & 61.5 & - &  60.6\\
\model{mPLUG-Owl2} \model{(8.2B)}~\cite{ye2023mplugowl2} & 79.4 & 56.1 & 57.7 & - & - & - & - & 54.3* & 86.2 &  64.5\\
\model{LLaVA-1.5} \model{(Vicuna-13B)}~\cite{liu2023improved} & 80.0 & 63.3 & - & - & - & 76.9* & 65.4* & 61.3* & 85.9 & 67.7\\
% \midrule
\hline
\rowcolor{lightgray}
\multicolumn{11}{l}{\textit{Our instruction-tuned baselines}} \\
\model{PaLI-3-Instruct} \model{(5B)} & 79.9 & 59.7 & 56.7 & 78.3 & 57.6 & 81.9 & 70.4 & 63.3* & 87.7 & 68.6 \\
\model{PaLI-X-Instruct} \model{(55B)} & 83.6 & 63.3 & 64.3 & 84.1 & 61.5 &  85.5 & 75.4 & 65.0* & \textbf{88.9} & 75.0 \\
\rowcolor{lightgray}
\multicolumn{11}{l}{\textit{Our visual program distillation models}} \\
\textbf{\model{PaLI-3-VPD} \model{(5B)}} & \underline{80.4} & \underline{61.3} & \underline{57.5} & \underline{78.5} & 56.5 & \underline{83.1} & \underline{70.9} & \underline{63.7*} & 
\underline{88.6} & \underline{69.0} \\

\textbf{\model{PaLI-X-VPD} \model{(55B)}} & \textbf{83.9} & \textbf{64.9} & \textbf{64.6} & \textbf{84.5} & \textbf{62.7} & \textbf{86.2} & \textbf{76.6} & \textbf{65.4*} & 88.8 & \textbf{76.2} \\
\bottomrule[1.2pt]
\end{tabular}
}
\vspace{-1ex}
\caption{
Comparison with SOTA generalist VLMs. \model{PaLI-X-VPD} outperforms all prior models. VPD improves performance on $8/9$ tasks for both \model{PaLI-3} and \model{PaLI-X}, and is particularly effective on counting questions (TallyQA), compostional questions (GQA), and the comprehensive benchmark (MMBench). \underline{Underline} indicates when \model{PaLI-3-VPD} outperforms the \model{Instruct} version. * marks tasks unseen during training.
POPE and MMBench are zero-shot benchmarks for all models.
}
% \vspace{-2ex}
\label{tab:generalist_scores}
% \bottomrule
\end{table*}

\subsection{Experimental setup}
\label{sec:experiment:setups}

\paragraph{Backbone models.}
We use two state-of-the-art VLMs, \model{PaLI-3} \model{(5B)}~\cite{chen2023pali3} and \model{PaLI-X} \model{(55B)}~\cite{chen2023palix} as our base models. Both take images and text as input, and generate text as output. % We include details about the model architectures and pre-training strategy in \S\ref{sec:appendix:training_details:pali_pretraining}.
For simplicity, we further refer to them as ``PaLI model'' when we discuss steps performed with each of them individually.

\paragraph{Data for Generalist Models.}
We fine-tune the pre-trained PaLI model on two types of datasets to make it a \textit{generalist} VLM—that is, a model that performs relatively well on any task without further training on that specific task:
% \begin{enumerate}[noitemsep,topsep=0pt,wide, labelwidth=!, labelindent=0pt]
% \item
% \item 

\textbf{(1)} \textit{Multimodal Instruction-Tuning (MMIT) tasks}, created in the spirit of Self-Instruct~\cite{wang2022selfinstruct}. An LLM is prompted with image captions, and generates task inputs and the desired outputs about the corresponding image. Details of MMIT tasks are covered in~\cite{wang2023nonintrusive}. 
    Note that these tasks cover a wide variety of common use cases, but do not include the specific in-domain tasks used in this work.
% \item 

\textbf{(2)} \textit{Academic task-oriented VQA tasks}. Since image captions contain only a coarse description of visual information in the image, and may miss details that are important for solving the task. Additionaly, LLM are likely to hallucinate during this data curation process. To further boost the accuracy of our VLMs, we also fine-tune the PaLI models with academic task-oriented VQA tasks. %(details and data statistics are shown in \S\ref{sec:appendix:training_details:data-mixing-academic}).
The data mixture covers subsets of a wide variety of VQA tasks, including general VQA (VQAv2~\cite{goyal2017making}), optical character recognition (OCRVQA~\cite{mishraICDAR19ocrvqa}), compositional questions and reasoning (GQA~\cite{hudson2019gqa}), counting (TallyQA~\cite{acharya2019tallyqa}), and VQA that involves external knowledge (OK-VQA~\cite{marino2019okvqa} and A-OKVQA~\cite{schwenk2022aokvqa}). The tasks contain a textual query and a short expected label. 
We use the pipeline in \S\ref{sec:method:generate_cot} to synthesize CoT reasoning steps for these labels, and tune the PaLI model with these the loss in  Equation~\ref{equation:loss}.
Notice that sometimes the pipeline fails to find a program that generates the correct answer. In that case, we set $\mathcal{L}_{rationale}$ to $0$ and only keep $\mathcal{L}_{label}$. \S\ref{sec:experiment:results} shows how many programs are kept after the filtering stage.
More details of the training data mixture is in \S\ref{sec:appendix:training_details}.
% \end{enumerate}

% The data mixture covers subsets of a wide variety of VQA tasks, including general VQA (VQAv2~\cite{goyal2017making}), VQA that involves OCR (OCRVQA~\cite{mishraICDAR19ocrvqa}), VQA that involves compositional questions and reasoning (GQA~\cite{hudson2019gqa}), VQA that involves external knowledge (OK-VQA~\cite{marino2019okvqa} and A-OKVQA~\cite{schwenk2022aokvqa}), and VQA about counting~\cite{acharya2019tallyqa}. 

\paragraph{Data for specialist models.} 
While fine-tuning the generalist model with VPD, we only use a subset of each task's training data.
%Also, each task has different annotation criterion.
To evaluate our model's ability on each individual task, we continue fine-tuning \model{PaLI-3-VPD} and \model{PaLI-X-VPD} on each individual task on the training splits.

\paragraph{Training setup.}
Both PaLI-3 and PaLI-X follow an encoder-decoder architecture, where images are encoded into visual tokens via a visual encoder and then passed into a UL2 model.
Due to resource constraint, we use LoRA~\cite{hu2021lora} to fine-tune both PaLI models.
Specifically, we add LoRA weights on each linear layer in the attention blocks and the MLP blocks for both the encoder and decoder in the UL2 transformer.
More training details are in \S\ref{sec:appendix:training_details}.

\paragraph{Evaluation setup.}
We evaluate our models on a wide range of tasks, including various VQA tasks and recent zero-shot VLM benchmarks.
%VQAv2~\cite{goyal2017making} targets general visual understanding. GQA~\cite{hudson2019gqa} contains visual question that are compositional and requires reasoning. OK-VQA~\cite{marino2019okvqa} and A-OKVQA~\cite{schwenk2022aokvqa} are VQA tasks that require external knowledge beyond the image to answer. 
Noted that A-OKVQA contains two kinds of questions, multiple-choice (MC) and direct answer (DA). We report results on both.
TallyQA~\cite{acharya2019tallyqa} contains complex counting questions that involve object relationships, attribute identification, and reasoning to get the correct answer. It contains two evaluation partitions, simple and complex.
TextVQA~\cite{singh2019towards} focuses reading texts. We include it to evaluate our models on zero-shot tasks.
In addition to VQA tasks, we also test our models on two popular VLM benchmarks.
POPE~\cite{li2023pope} focuses on VLM hallucination, containing binary questions of whether or not an object exists in the image.
MMBench~\cite{liu2023mmbench} is a robust and comprehensive VLM benchmark testing a range of fine-grained abilities (e.g., object localization, attribute recognition, spatial relationship).
Details of the evaluation sets and metrics are in \S\ref{sec:appendix:training_details}. 

\paragraph{Baselines.}
We refer to the two PaLI models fine-tuned with VPD as \model{PaLI-3-VPD} and \model{PaLI-X-VPD}, and compare them with various baselines. To evaluate the effectiveness of VPD, we experiment with removing the synthesized CoT from our data mixture, and train the PaLI models with the exact same hyper-parameters and steps.
We call these models as 
\model{PaLI-3} \model{Instruct} and \model{PaLI-X} \model{Instruct}. For a fair comparison, these models are also trained with the same supervised loss for predicting the ground truth answers, on the same images and textual queries as our VPD variant.
Moreover, we also compare with the most recent SOTA vision-language models. %, including \model{Flamingo}~\cite{alayrac2022flamingo},  \model{MiniGPT-4}~\cite{zhu2023minigpt4}, \model{InstructBLIP}~\cite{dai2023instructblip}, \model{Shikra}~\cite{chen2023shikra}, \model{Qwen-VL}~\cite{bai2023qwenvl}, \model{mPLUG-Owl2}~\cite{ye2023mplugowl2}, \model{LLaVA 1.5}~\cite{liu2023improved}, and \model{CogVLM}~\cite{wang2023cogvlm}.
These VLMs are initialized with pre-trained visual encoders and LLMs, and then trained with image-text pairs, LLM-generated data, and academic tasks.

\subsection{Quantitative results}
\label{sec:experiment:results}

\vspace{-1ex}

\paragraph{Generalist model.}
Table~\ref{tab:generalist_scores} compares VPD with the baselines discussed in \S\ref{sec:experiment:setups}. % on VQAv2, GQA, OK-VQA, A-OKVQA (Multiple-choice and Direct Answer), TallyQA (Simple and Complex), TextVQA, POPE, and MMBench. 
We infer all answers by open-ended generation with the prompt \prompt{Answer with a single word or phrase.}, using greedy decoding without any constraint on the model's output space.
For multiple-choice questions, we run inference with the prompt \prompt{Answer with the option letter from the given choices directly.} and generate the option letter.
As Table~\ref{tab:generalist_scores} shows, \textbf{\model{PaLI-X-VPD} sets the new state-of-the-art on all benchmarks}. Compared with prior generalist VLMs, it achieves significant improvement on MMBench ($+8.5$), TallyQA complex ($+9.8$), and A-OKVQA($+9.5$ compared with specialist SOTA\cite{dai2023instructblip}).
\model{PaLI-3-VPD}, despite its small architecture size (5B), outperforms all prior models on VQAv2, TallyQA, POPE, and MMBench, including much larger models that use Vicuna-13B as backbones~\cite{dai2023instructblip, liu2023improved, chen2023shikra}. However, its performance on knowledge-based VQA tasks does not outperform the prior SOTA. Our hypothesis for this is that its language model (3B UL2) is too small to contain all the necessary knowledge needed for correctly solving these tasks.

When compared to their \model{Instruct} variants, both \model{PaLI-3-VPD} and \model{PaLI-X-VPD} outperform on $11/12$ tasks. 
Specifically, for \model{PaLI-X}, VPD obtains a $+1.6$ improvement on GQA (which is heavily focused on compositional questions, spatial relationship, and localization), $+1.2$ on TallyQA for complex questions, and $+1.2$ on MMBench. These results suggest that \textbf{VPD is a more effective method for creating instruction-tuning data which enables VLMs to improve their visual reasoning ability}.

\begin{table*}[h]
\small
\centering
% \vspace{-1ex}
{\fontsize{8.5pt}{9.5pt}\selectfont
\begin{tabular}{lcccccccc}
\toprule[1.2pt]
& GQA & OK-VQA& \multicolumn{4}{c}{A-OKVQA} & \multicolumn{2}{c}{TallyQA} \\
Specialist Models &  &  & \multicolumn{2}{c}{Multi-choice} & \multicolumn{2}{c}{Direct Answer} & Simple & Complex \\
& test-dev & val & val & test & val & test & test & test \\
\midrule
\model{InstructBLIP (Vicuna-7B)}~\cite{dai2023instructblip} & - & 62.1 & 75.7 & 73.4 & 64.0 & 62.1 & - & -\\
\model{PaLI-3 (5B)}~\cite{chen2023pali3} & - & 60.1 & - & - & - & - & 83.3 & 70.5 \\
\model{PaLI (17B)}~\cite{chen2022pali} & - & 64.5 & - & - & - & - & 81.7 & 70.9\\
\model{CogVLM (17B)}~\cite{wang2023cogvlm} & 65.2 & 64.7 & - & - & - & - & - & - \\
\model{PaLI-X (55B)}~\cite{chen2023palix} & - & 66.1 & - & - & - & - & 86.0 & 75.6\\
\midrule
\model{PaLI-3-VPD} (5B) \textit{generalist} & 61.3 & 57.5 & 78.5 & - & 56.5 & - & 83.1 & 70.9 \\
\model{PaLI-3-VPD} (5B) \textit{specialist} & 64.7 & 60.3 & 79.7 & 76.5 & 65.5 & 63.6 & 83.3 & 70.8\\
\midrule
\model{PaLI-X-VPD} (55B) \textit{generalist} & 64.9 & 64.6 & 84.5 & - & 62.7 & - & \textbf{86.2} & \textbf{76.6} \\
\model{PaLI-X-VPD} (55B) \textit{specialist} & \textbf{67.3} & \textbf{66.8} & \textbf{85.2} & \textbf{80.4} & \textbf{71.1} & \textbf{68.2} & \textbf{86.2} & 76.4\\
\bottomrule[1.2pt]
\end{tabular}
}
% \vspace{-1ex}
\caption{
Comparison of per-task fine-tuning results of specialist models.  \model{PaLI-X-VPD} sets a new SOTA for all the tasks. %outperforming the prior SOTAs set by \model{PaLI-X} and other VLMs.
}
\label{tab:pertask_scores}
% \bottomrule
\end{table*}

% PaLI-3 (5B)~\cite{chen2023pali3} and PaLI-X (55B)~\cite{chen2023palix}

\paragraph{Per-task fine-tuning (specialist).}
The results for the specialist models are shown in Table~\ref{tab:pertask_scores}. 
\textbf{\model{PaLI-X-VPD} sets a new SOTA on all benchmarks}.
Note how the specialist models tend to have higher scores than the generalist one. We propose several hypotheses for this performance gain:
(1) As shown in Table~\ref{tab:pertask_scores}, the score gap tends to be larger on free-form VQA tasks. This may be due the fact that human annotations on these datasets have ambiguities. 
For example, for the question \prompt{Who is looking up?}, GQA~\cite{hudson2019gqa} labels are \prompt{man} or \prompt{woman} while OK-VQA~\cite{marino2019okvqa} have more detailed labels, for example, \prompt{worker} or \prompt{cook}. 
Per-task fine-tuning alleviates this annotation ambiguity and lets the model focus on the annotation style of that task.
For multiple-choice and counting tasks, the answer has less ambiguity, and the score gaps are much smaller.
(2) The performance gap between \model{PaLI-3} \model{(5B)} \textit{specialist} and \textit{generalist} is larger than that of \model{PaLI-X}. We hypothesize that this shows models with larger scale has more multi-task capacity;
(3) Per-task fine-tuning adds more in-domain training data, which generally improves performance.

% \paragraph{Is chain-of-thought useful during inference?}
% \input{tables/cot_vs_label}

\begin{figure}[h!]
% \vspace{-1ex}
\centering
  \includegraphics[width=\linewidth]{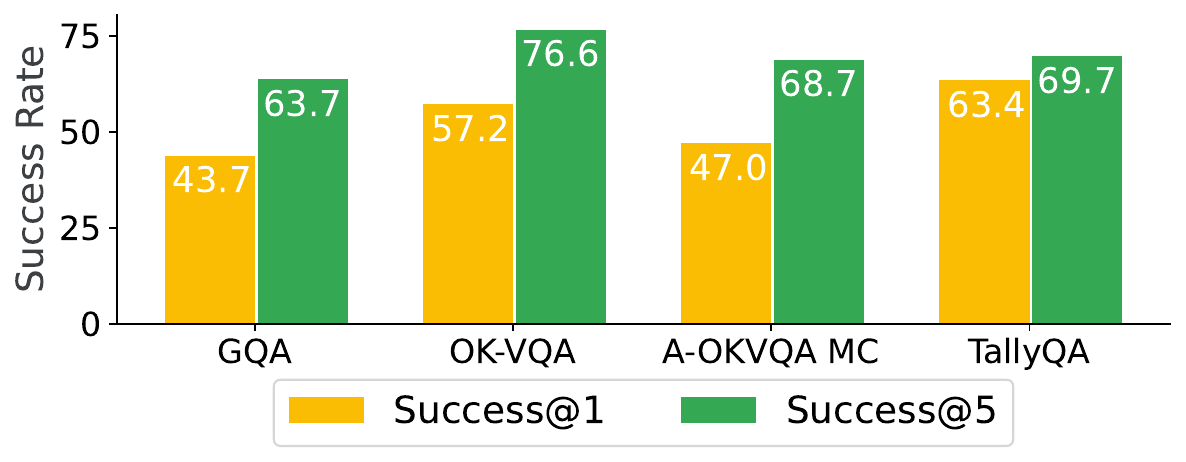}
   \vspace{-3mm}
  \caption{
Success rate of top-$1$ program and top-$5$ programs on the training set during our data synthesis process.
}
\vspace{-2mm}
  \label{fig:top1_vs_top5}
\end{figure}

\begin{figure}[h!]
\centering
  \includegraphics[width=\linewidth]{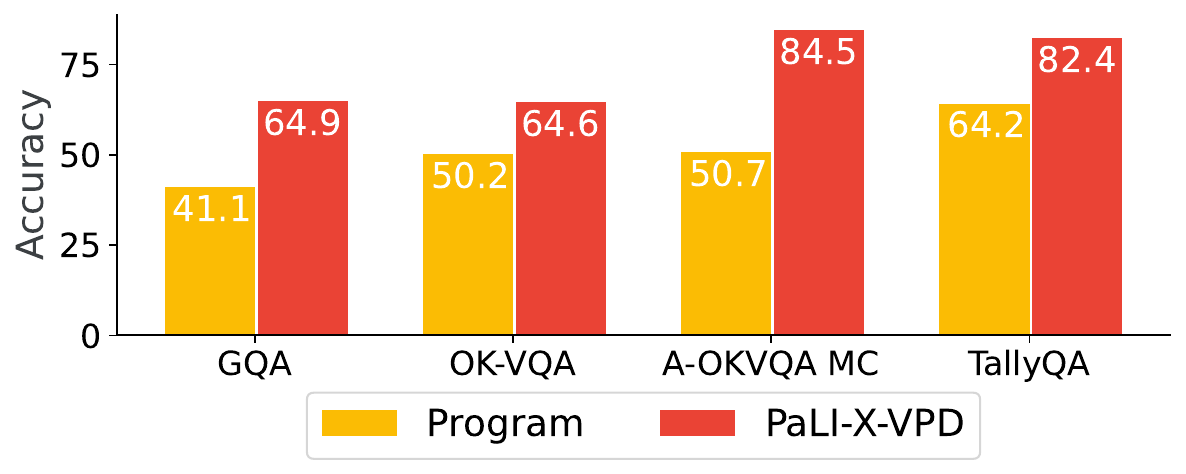}
  \vspace{-3ex}
  \caption{
Accuracy of visual programs and \model{PaLI-X-VPD} on validation sets.
}
  \label{fig:program_vs_vpd}
  \vspace{-3ex}
\end{figure}

\paragraph{Analysis: sampling multiple programs is key to good data generation.}
Fig.~\ref{fig:top1_vs_top5} shows the success rate of finding at least one program that passes the filtering stage, when the LLM generates the top-$1$ or top-$5$ programs, respectively. There is a dramatic increase in success rate from $1$ program to $5$: $+45\%$ on GQA and A-OKVQA, $+33\%$ on OK-VQA, and $+10\%$ on TallyQA. This design choice greatly improves our data synthesis efficiency, and, as a consequence, adding more CoT data that requires complex reasoning in our training set.
We also conduct an analysis in \S\ref{sec:appendix:comparison_with_programs} to compare the performance of VPD models with that of the visual programs they are distilled from.

\paragraph{Analysis: comparison with visual programs}
\label{sec:appendix:comparison_with_programs}
Figure~\ref{fig:program_vs_vpd} does a side-by-side comparison of the accuracy of visual programs and that of \model{PaLI-X-VPD} on GQA, OK-VQA, A-OKVQA (multiple choice), and TallyQA (simple and complex combined). 
We report results on the validation sets, so \model{PaLI-X-VPD} was not distilled with these exact visual programs, but with visual programs generated in a similar manner on the training set.
The results indicate that \model{PaLI-X-VPD} has much higher accuracy than visual programs on all tasks. This raises an interesting question: why is the student model more accurate than its teacher? 
% One potential reason is that visual programs, as a neural-symbolic method, can only be used in a zero-shot fashion, while VLMs can be trained in an end-to-end fashion. \otilia{This argument seems incomplete, how does this lead to higher accuracy?}
One explanation is that our pipeline allows us to leverage labeled data to improve the quality of the visual programs. When ground truth labels are available, we can choose a correct program among $5$ candidates, rather than only relying on a single candidate. As supported by the results in Figure~\ref{fig:top1_vs_top5}, this greatly improves the accuracy of our visual programs as the teacher, thus making them more helpful for distillation.

\begin{figure*}[th!]
\vspace{-1ex}
\centering
  \includegraphics[width=0.9\linewidth]{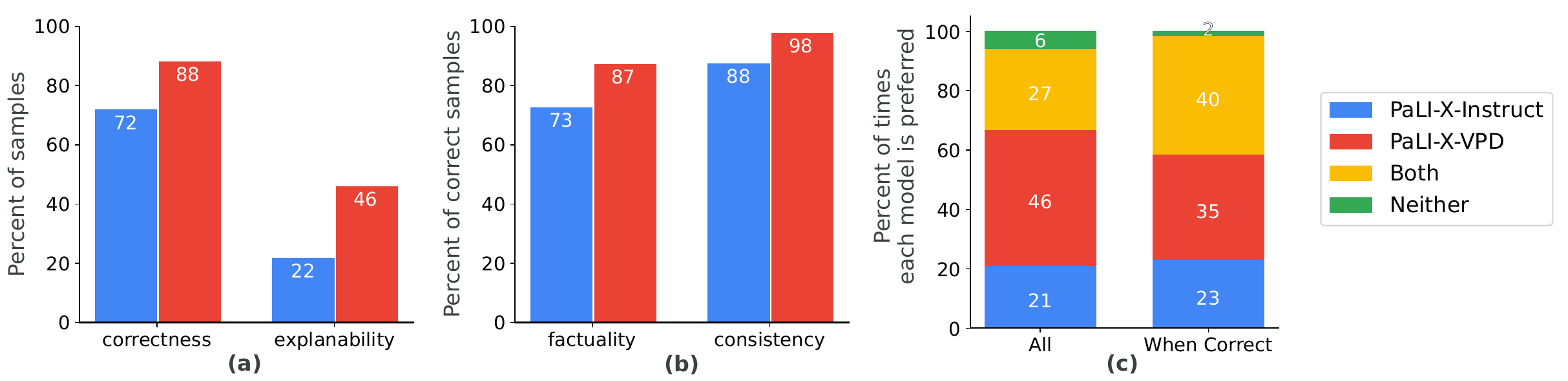}
  \vspace{-1ex}
  \caption{
Human evaluation results assessing answer and rational quality for \model{PaLI-X} \model{Instruct} and \model{PaLI-X VPD}: \textbf{(a)} percentage of answers that are correct and have explanations; \textbf{(b)} rationale factuality and consistency for the answers that contain an explanation; \textbf{(c)} preference between the two models, when aggregating across all samples (``All'') and across those with correct answers (``When correct'').
% \ranjay{These colors are fine. Can we increase font size in figures 4,5,6,7 so that the smallest font is the caption font size?}
}
  \label{fig:human_eval}
\end{figure*}

\subsection{Human evaluation on rationales}
\label{sec:experiment:human_eval}

In this section, we focus on the quality of model outputs. We performed this analysis using human annotators, who are asked to evaluate both the correctness of the final answer and the quality of the rationale behind it.

\paragraph{Models.}
We compare \model{PaLI-X-VPD} with \model{PaLI-X} \model{Instruct}.
Among possible baselines, we chose \model{PaLI-X} \model{Instruct} because it is trained to generate long-form answers~\cite{wang2023nonintrusive}, which allows us to assess if it is the quality of \model{PaLI-X-VPD}'s answers that the annotators prefer, rather than its length. Since \model{PaLI-X} \model{Instruct} is also instruction-tuned, it can be prompted to provide long answers to alleviate this confounder.
% We prompted \model{PaLI-X-VPD} with \prompt{Explain the rationale to answer the question}, and \model{PaLI-X} \model{Instruct} with \prompt{Explain in detail}. 

\paragraph{Annotation protocol.} We run inference with each of the two models on a combination of $600$ samples from GQA (test-dev), A-OKVQA (val), and TallyQA (Simple and Complex) and record their answers and rationales. We then ask 3 human annotators to evaluate each model answer.
We use prior work from natural language processing (NLP) as inspiration, and build upon it for selecting the evaluation criteria~\cite{ye2022unreliability}.
Given an image and a query, for each model-generated rationale, we ask human annotators to score the model answers along the following criteria:
(1) \textbf{correctness}—is the final answer correct?
(2) \textbf{explainability}—does the model explain its rationale for reaching the final answer?
(3) \textbf{factuality}—is every step in the rationale factually correct (with respect to the image and external knowledge)?
(4) \textbf{consistency}—does step and final answer logically follow from the previous ones?
Note that a rationale may have the wrong answer while being consistent.
We also conduct a \textbf{side-by-side comparison}, and ask annotators which of the two answers—the one provided by \model{PaLI-X-VPD} or by \model{PaLI-X} \model{Instruct}—they prefer in terms of quality of the answer and explanation.
More details about the annotation protocol are in \S\ref{sec:appendix:human_annotation}.

\paragraph{Human evaluation results.} The results of the evaluation are first averaged among the human raters per sample, then aggregated across samples.
\textbf{Our \model{PaLI-X-VPD} model far outperforms \model{PaLI-X} \model{Instruct} along all criteria}.
Fig.~\ref{fig:human_eval} (a) shows the correctness of the final answer provided by each model (i.e. accuracy), and the proportion of the samples where the model provides a CoT rationale to explain its answer.
\model{PaLI-X-VPD}'s gain in accuracy of $+16.7\%$ is even more impressive than in evaluations in \S\ref{sec:experiment:results} based on benchmark labels—this is because human annotators are better able to assess correctness for ambiguous questions with different possible interpretations (e.g., the model's answer \prompt{In the living room.} to the question \prompt{Where is the couch located?} was considered correct by annotators, even when the benchmark answer was \prompt{On the right.}).
Moreover, the explainability results confirm that our model is able to explain its own answer on +24\% more samples than the instruction tuned model.
Additionally, Fig.~\ref{fig:human_eval} (b) shows \textbf{impressive rationale quality}: among the samples where an explanation is provided, \model{PaLI-X-VPD}'s rationales are \textbf{factual 87.2\% of times}, and \textbf{consistent 97.8\% of times}, a gain of $+14.6\%$ and $+10\%$, respectively, when compared to \model{PaLI-X} \model{Instruct}.
% \model{PaLI-X Instruct}'s rationales are factual \red{$72.6\%$} of times, and consistent \red{$87\%$} times, whereas

We also asked the annotators which of the two answers given by the two models they prefer.
% We observed that there are cases where one or both models have similar answers, or both answers are incorrect so it would be difficult to choose a favorite, so we allowed annotators to also choose ``Both'' or ``Neither''.
We aggregated these results in two ways: (1) across all samples, (2) across the samples where both models answered the question correctly. The results in Fig.~\ref{fig:human_eval} (c) show that \model{PaLI-X-VPD} is preferred on $25\%$ more samples than \model{PaLI-X} \model{Instruct} in the ``All'' case, and on $12\%$ more samples when both are correct. This suggests that even when \model{PaLI-X-VPD} makes mistakes, it still provides a better answer. Such examples are shown in \S\ref{sec:appendix:human_annotation}.
These results confirm that a model fine-tuned with VPD leads to more faithful and consistent answers.

% \subsection{Comparison with Visual Programs}
% \label{sec:experiment:analysis}

% \begin{figure}[th!]
% \centering
%   \includegraphics[width=\linewidth]{figures/program_top1_vs_top5.pdf}
%   \vspace{-7mm}
%   \caption{
% Success rate of top-$1$ program and top-$5$ programs on the training set during our data synthesis process.
% }
 
%   \label{fig:top1_vs_top5}
% \end{figure}

% \paragraph{Sampling multiple programs is key to good data generation.}
% Figure~\ref{fig:top1_vs_top5} shows the success rate of finding at least one program that passes the filtering stage, when the LLM generates the top-$1$ or top-$5$ programs, respectively. There is a dramatic increase in success rate from $1$ program to $5$: $+45\%$ on GQA and A-OKVQA, $+33\%$ on OK-VQA, and $+10\%$ on TallyQA. This design choice greatly improves our data synthesis efficiency, and, as a consequence, adding more CoT data that requires complex reasoning in our training set.

% We also conduct an analysis in Appendix~\ref{sec:appendix:comparison_with_programs} to compare the performance of VPD models with that of the visual programs they are distilled from.

\section{Experiments on content moderation}
\label{sec:exp:content}

Prior experiments focus on training generalist VLMs.
Here, we explore the effectiveness of VPD at quickly adapting models to real-world applications from a different domain than the training data~\cite[e.g.,][]{stretcu2023agile}. 
We experiment on Hateful Memes~\cite{kiela2020hateful}, a content moderation dataset where the task is to classify if a meme contains hateful content. The target labels are ``yes'' or ``no'', and models are evaluated in terms of classification accuracy and AUC-ROC. % (examples in \S\ref{sec:appendix:examples}).
We experiment with two settings: \textit{supervised}, in which the models are trained on the provided training set with $8,500$ labels, and \textit{unsupervised}, in which no human labels are provided. We provide qualitative examples of the rationales our models give in Figure~\ref{fig:hateful_memes_example}.
% The results are in Table~\ref{tab:hateful_memes}.

\begin{table}[h]
\vspace{-1ex}
\small
\centering
% \begin{tabular}{P{1.7cm}P{1.2cm}P{1.2cm}P{1.0cm}P{1.2cm}P{1.2cm}P{1.2cm}P{1.1cm}P{1.1cm}P{1.1cm}}
{\fontsize{8pt}{9.5pt}\selectfont
\begin{tabular}{l|cc}
\toprule[1.2pt]
Model & Acc & AUC-ROC \\
% \midrule
\hline
\rowcolor{lightgray}
\multicolumn{3}{l}{\textit{Unsupervised / Zero-Shot Methods}}\\
% \model{Flamingo} \model{(9B)}~\cite{alayrac2022flamingo} & 57.0 & - \\
% \model{InstructBLIP} \model{(Vicuna-13B)}~\cite{dai2023instructblip} & 59.6 & -\\
% \model{MiniGPT-V2} \model{(7B)}~\cite{chen2023minigptv2} & 57.8 & -\\
% % \midrule
% \hline
\model{Generated Programs} \textbf{(ours)} & 69.7 & 70.1 \\
% \model{PaLI-X-VPD} \model{(generalist)} & 61.4 & 66.8 \\
%\textbf{\model{PaLI-X-VPD} (specialist w/ 0-shot CoT)} & \textbf{70.8} & \textbf{78.3} \\
\hline
\rowcolor{lightgray}
\multicolumn{3}{l}{\textit{Supervised Methods}}\\
\model{VisualBERT}~\cite{Li2019VisualBERTAS} & 69.5 & 75.4 \\
\model{Flamingo (80B)}~\cite{alayrac2022flamingo} & - & 86.6 \\ 
\model{Previous SOTA}~\cite{mei2023improving} & 78.8 & 86.7 \\
\model{PaLI-X-VPD} \model{(label-only FT)} & 77.6 & 88.0 \\
\textbf{\model{PaLI-X-VPD} (specialist w/ CoT)} & \textbf{80.8} & \textbf{89.2} \\
\hline
\rowcolor{lightgray}{\model{Human}}~\cite{kiela2020hateful} & 84.7 & 82.7 \\
% \textcolor{shadecolor}{\model{Human}}~\cite{kiela2020hateful} & \textcolor{shadecolor}{84.7} & \textcolor{shadecolor}{82.7} \\
\bottomrule[1.2pt]
\end{tabular}
}
\vspace{-1ex}
\caption{
Results on Hateful Memes~\cite{kiela2020hateful} seen test set. We improve SOTA for both zero-shot and supervised settings. 
% Training with program-generated CoTs allows unsupervised \model{PaLI-X-VPD} to outperform strong supervised baselines.
Unsupervised \model{PaLI-X-VPD} outperforms strong supervised baselines.
}
\label{tab:hateful_memes}
% \bottomrule
\vspace{-3mm}
\end{table}

\vspace{-1ex}

\subsection{Supervised setting}
We first experiment with the supervised setting. We fine-tune the models on 8,500 labels in two ways:
\begin{enumerate}[noitemsep,topsep=0pt]
    \item \uline{Label-only fine-tuning}: To establish a strong baseline, we first experiment with the traditional supervised setting, where we fine-tune the model to output ``yes'' or ``no''.
    \item \uline{PaLI-X-VPD (specialist)}: We use the complete VPD pipeline to train this model. We select an execution trace that leads to the correct label, and use it to tune our VLM. 
\end{enumerate}

\paragraph{Results.}
Our \model{PaLI-X-VPD} \model{(specialist)} \textbf{sets a new SOTA for this task}, with an accuracy of $80.8\%$ and AUC-ROC of $89.2\%$. 
% As shown in Figure~\ref{fig:hateful_memes_example} (c), \model{PaLI-X-VPD} (supervised specialist) is able to capture the nuances in the meme, while unsupervised methods fail.
It outperforms the label-only fine-tuning baseline and significantly improves the SOTA metrics, \textbf{achieving nearly human-level accuracy, and super-human AUC-ROC}.
% Despite the success of our VPD models, we still observe some failure cases like (d), in which even our supervised model fails while the program succeeds on hateful content detection.

\subsection{Unsupervised setting}
How would VPD perform if we do not have any human-annotated labels? We experiment with three methods:
\begin{enumerate}[noitemsep,topsep=0pt]
    \item \uline{Generated Program}: The program generated by PaLM-2~\cite{anil2023palm2} for solving this task consists of following main steps: (1) get image description with PaLI-X~\cite{chen2023palix}; (2) use an OCR tool to extract embedded texts; (3) given the image description and OCR texts, ask PaLM-2 to explain if this meme is hateful.
    \item \uline{PaLI-X-VPD (generalist)}: In a zero-shot setting, we directly prompt our \model{PaLI-X-VPD}\model{(generalist)} with: \prompt{The text is <OCR text>. Is this a hateful meme?} We compute the probability of \model{PaLI-X-VPD} generating ``yes'' or ``no'' and measure accuracy and AUC-ROC.
    \item \uline{PaLI-X-VPD (specialist with zero-shot CoT)}: We follow our VPD pipeline and convert the execution traces of the program into CoTs, and then fine-tune \model{PaLI-X-VPD} to output these CoTs. Since no groundtruth labels are available, no filtering is done during the process.
    % Here we use VPD to distill the execution traces of our generated programs into \model{PaLI-X-VPD}. Since have no labels labels, the programs used in VPD are only filtered against failure to execute, but are not verified for correctness. %We execute the selected program and use it to synthesize a CoT.
\end{enumerate}

\begin{figure*}[h]
\centering

  \includegraphics[width=0.75\linewidth]{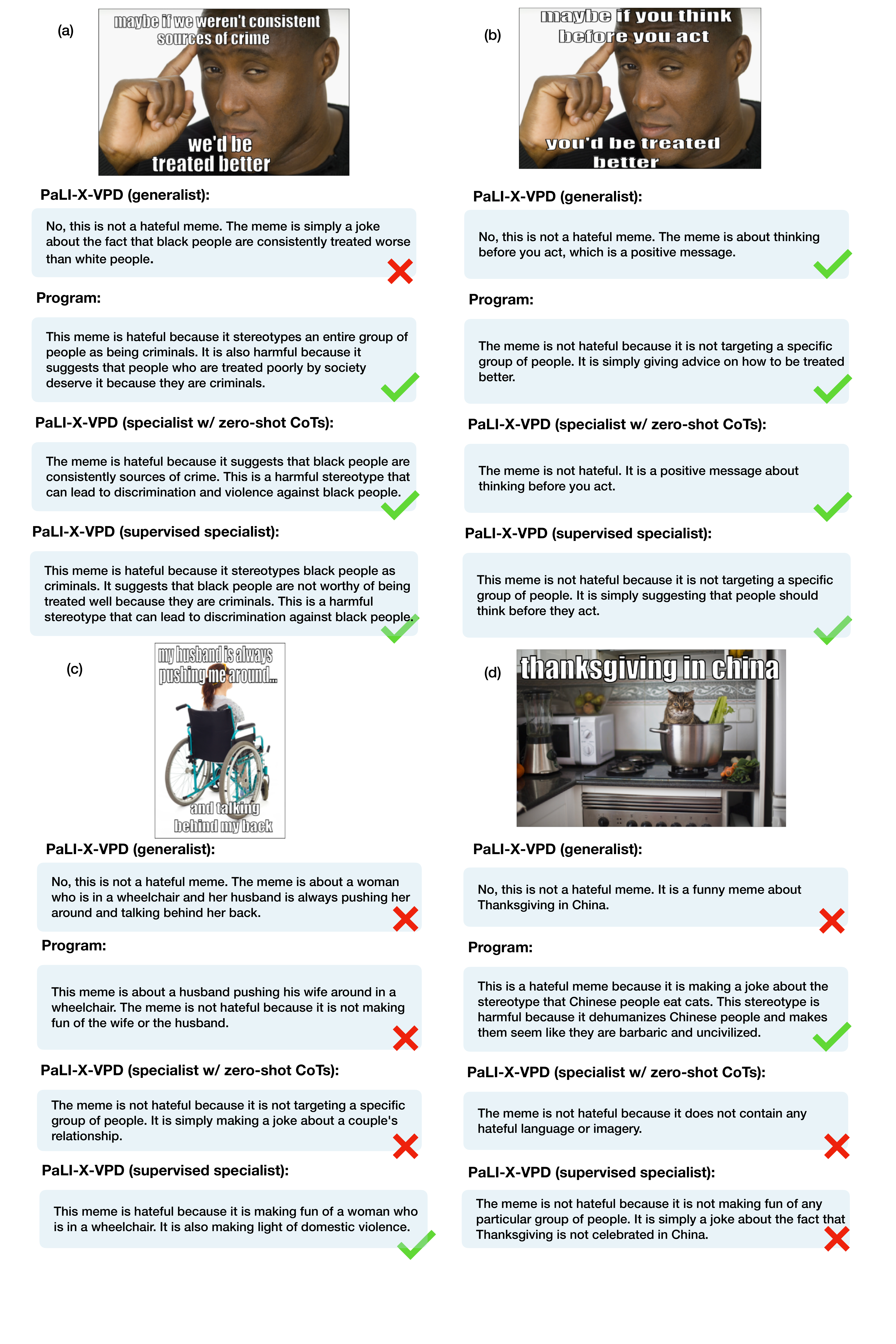}

  \caption{
Example outputs of different methods on Hateful Memes~\cite{kiela2020hateful} dev set. Here we prompt our \model{PaLI-X-VPD} models with ``Is this a hateful meme? Explain the rationale to answer the question.'' to get models' long-form rationales. The unsupervised methods include zero-shot \model{PaLI-X-VPD (generalist)}, our generated program, and \model{PaLI-X-VPD} (specialist with zero-shot CoTs). We also include our supervised method, i.e., \model{PaLI-X-VPD} (specialist). We mark whether their outputs match the gold answers.
}
  \label{fig:hateful_memes_example}
\end{figure*}

\paragraph{VPD significantly improves performance even when no labels are available.}
As shown in Table~\ref{tab:hateful_memes}, \model{PaLI-X-VPD (generalist)} outperforms all other VLMs ($61.4\%$).
Interestingly, the generated programs themselves get much higher accuracy ($69.7\%$) than on the previous datasets, perhaps because PaLM-2 is better suited at analyzing a meme than typical VQA datasets.
As shown in Figure~\ref{fig:hateful_memes_example}, \model{PaLI-X-VPD (generalist)} is relatively insensitive to hateful content, while programs are better.
Moreover, our \model{PaLI-X-VPD} \model{(specialist)} sets a new zero-shot SOTA on Hateful Memes, and even outperforms supervised VisualBERT~\cite{Li2019VisualBERTAS}.
To understand this impressive gain, we manually inspect the model outputs and find that the model has learnt PaLM-2's reasoning process via VPD, as seen in Figure~\ref{fig:hateful_memes_example}.
Besides, our model has better understanding of the image. As seen in Figure~\ref{fig:hateful_memes_example} (a), the program, despite giving the correct answer, has not mentioned anything about ``black people'', because it is not covered in the image caption. In comparison, our \model{PaLI-X-VPD (specialist)} includes it and gives a better explanation.

\paragraph{More qualitative analysis.} We observe that \model{PaLI-X-VPD} (supervised specialist) is able to capture the nuances in the meme, while unsupervised methods fail, as shown in Figure~\ref{fig:hateful_memes_example} (c). Also, there are some failure cases like (d), in which even our supervised model fails while the program succeeds on hateful content detection.

% To understand this impressive gain, we manually inspect the model outputs and find that the model has learnt PaLM-2's reasoning process via VPD. We provide a qualitative example in \S\ref{sec:appendix:examples}.

\section{Limitations and Future Directions}
\label{sec:appendix:future_work}

We mainly identify two directions for improvement. First, VPD can be further scaled-up with more diverse tasks. Second, we find that there are some problems that our visual program framework (ViperGPT~\cite{surismenon2023vipergpt}) cannot solve. Further improving the programs will yield bigger gains for VPD.
We list the limitations below, along with future work that may address these challenges.

\paragraph{Scaling-up VPD with LLM-generated questions and better filtering strategies.}  Our current setting limits the data source of VPD to existing VQA tasks. Future work can scale-up and diversify the tasks using LLMs, as demonstrated in Self-Instruct~\cite{wang2022selfinstruct}. One challenge is that there lacks human-annotated labels for LLM-generated tasks. Experiments in \S\ref{sec:exp:content} show that VPD works well even without labels. Nevertheless, it would be better if we have some filtering strategies, as shown in \S\ref{sec:experiment:results}. Since even the strongest VLMs like PaLI-X~\cite{chen2023palix} and GPT-4V~\cite{openai2023gpt4} cannot give reliable answers,  fact-checking strategies for multimodal chain-of-thought data is a promising future direction.

\paragraph{Agents, rather than static programs.}
There exist complex visual-language tasks that cannot be easily solved with one program. Recent work~\cite{yao2022react, Wu2023AutoGenEN, yang2023mmreact} have explored the idea of LLM agents, where LLMs interact with an environment and do planning interactively. We may be able to leverage this idea in our scenario, and have an LLM update its plans given the new information gathered from vision tools. % Future work can explore with agents, rather than one static program.

\paragraph{Adding fine-grained and dense labeling tools.} We find that programs struggles in complex scenarios when objects are occluded, or the scene contains too many objects. Adding dense-labeling tools like Segment Anything~\cite{alex2023segment} may address this challenge. For example, recently LISA~\cite{lai2023lisa} have proposed combining segmentation with LLMs. Future work can make dense labeling tools available in VPD in a similar way, which will further boost VLM performance.

\section{Conclusion}
We introduce VPD, a framework for distilling the reasoning abilities of LLMs along with the capabilities of vision tools into VLMs. VPD synthesizes training data for VLMs by generating programs that can leverage external tools. We use this technique to fine-tune some of the best existing VLMs (PaLI-3 and PaLI-X) and established new SOTA results on 8 classical VQA and 2 zero-shot multimodal benchmarks. According to human evaluations, VPD-tuned models provide more accurate answers and better rationales. Experiments on the Hateful Memes show how VPD can also adapt models to new real-world domains, even when no labeled data is available, also establishing new SOTAs.
We also point out directions to further improve VPD. We believe VPD will grant future VLMs better multimodal reasoning abilities.

\section{Acknowledgements}

We thank Tanmaya Dabral, Jialin Wu, Radu Soricut,  Aditya Avinash, Howard Zhou, Jiao Sun, Deqing Fu, Cyrus Rashtchian, Andrew Tomkins, Tom Duerig, as well as the Agile Modeling team for helpful project discussions and technical support in this project.
\newpage

{
    \small
    \bibliographystyle{ieeenat_fullname}
    \bibliography{main}
}

% \vspace*{0.5\textheight}
% \newpage

\onecolumn
\appendix


\section{Examples outputs of our data-synthesis pipeline}
\label{sec:appendix:examples}
We first present step-by-step examples of our data-synthesis pipeline. As discussed in \S\ref{sec:method}, the pipeline contains four steps: program generation, program execution, program filtering, and converting execution traces into chains-of-thoughts. Here we show three filtered examples from three datasets we use.
As shown, LLMs like PaLM-2~\cite{anil2023palm2} are able to generate complex programs to solve compositional vision tasks. The synthesized CoTs are used to fine-tune our VLMs.

\begin{figure*}[h]
\centering
  \includegraphics[width=\textwidth]{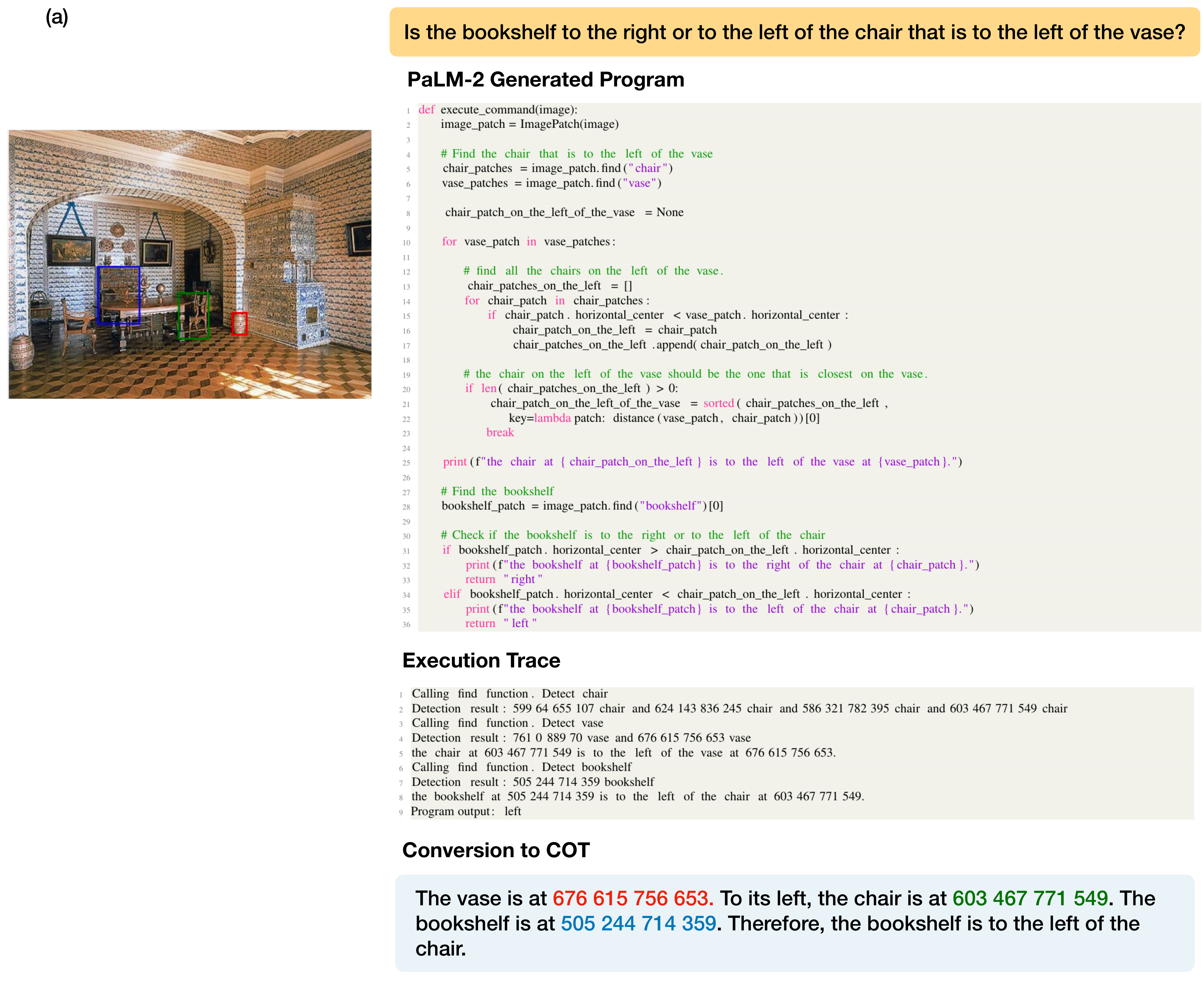}
  \vspace{-2ex}
\end{figure*}

\begin{figure*}[h]
\centering
  \includegraphics[width=\textwidth]{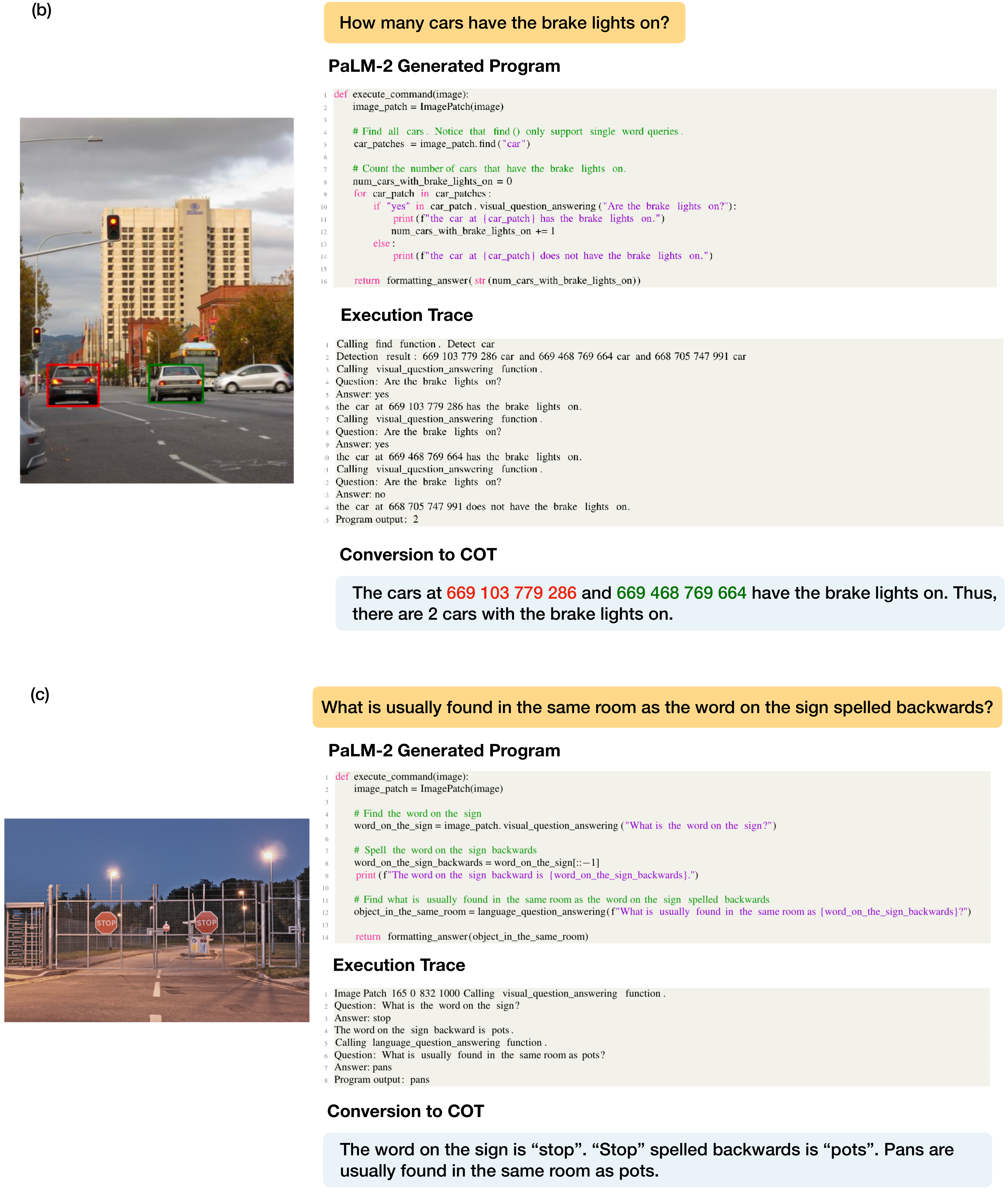}
%   \vspace{-2ex}
  \caption{Examples of our data-synthesis pipeline. (a) is from GQA~\cite{hudson2019gqa}; (b) is from TallyQA~\cite{acharya2019tallyqa}; (c) is from A-OKVQA~\cite{schwenk2022aokvqa}.
  }
\end{figure*}

\newpage

\section{Experimental details}
\label{sec:appendix:training_details}

\subsection{Model architecture details}
\label{sec:appendix:training_details:pali_pretraining}
For both PaLI models that we considered in our experiments, the architectures are similar: images are encoded into visual tokens individually via a visual encoder. Then, the vision tokens along with the textual inputs are passed through an encoder-decoder UL2 Transformer~\cite{tay2022ul2}. The PaLI models were then pre-trained with image-text pairs to perform multimodal tasks. Specifically, PaLI-3~\cite{chen2023pali3} uses a pre-trained 2B SigLIP~\cite{zhai2023siglip} as visual encoder, and a 3B UL2. The image resolution is $812 \times 812$.
PaLI-X~\cite{chen2023palix} uses a pre-trained VIT-22B~\cite{dehghani2023vit22b} as visual encoder, and a 32B UL2. The image resolution is $756 \times 756$. Please refer to the PaLI-3~\cite{chen2023pali3} and PaLI-X~\cite{chen2023palix} papers about more architecture details.

\subsection{Datasets}
The details of the data mixture of academic task-oriented VQA datasets used in VPD training are shown in Table~\ref{tab:data_mixture}.
We only use a subset of each dataset's training set. \# labels refers to the total number of examples (containing image, query, and answer) we use. \# CoTs refers to the number of examples that we have synthesized CoTs using our programs. In total, there are $89.6$K CoTs used during training.

\begin{table}[h]
\small
\centering
% \begin{tabular}{P{1.7cm}P{1.2cm}P{1.2cm}P{1.0cm}P{1.2cm}P{1.2cm}P{1.2cm}P{1.1cm}P{1.1cm}P{1.1cm}}
\begin{tabular}{l|l|rr}
\toprule[1.2pt]
Dataset & Description &\# labels & \# COTs \\
\midrule
VQAv2~\cite{goyal2017making} & General & 100.0K & \\
OCR-VQA~\cite{mishraICDAR19ocrvqa} & OCR & 50.0K & \\
GQA~\cite{hudson2019gqa} & Compositional & 86.0K & 38.0K \\
OK-VQA~\cite{marino2019okvqa} & Knowledge & 9.0K & 6.7K \\
A-OKVQA~\cite{schwenk2022aokvqa} & Knowledge & 17.1K & 11.2K \\
TallyQA~\cite{acharya2019tallyqa} & Counting & 48.4K & 33.7K \\
\midrule
Total & & 310.5K & 89.6K \\
\bottomrule[1.2pt]
\end{tabular}
% \vspace{-1ex}
\caption{
Data mixture of academic task-oriented VQA datasets used in VPD training.
}
\label{tab:data_mixture}
% \bottomrule
% \vspace{-3mm}
\end{table}

Details of each evaluation benchmark we use are in Table~\ref{tab:evaluation_data}.
For free-form question answering, we run inference with the prompt \prompt{Answer with a single word or phrase.}, using greedy decoding without any constraint on the model's output space.
For multiple-choice questions, we run inference with the prompt \prompt{Answer with the option letter from the given choices directly.} and generate the option letter.
\begin{table*}[ht]
\small
\centering
% \begin{tabular}{P{1.7cm}P{1.2cm}P{1.2cm}P{1.0cm}P{1.2cm}P{1.2cm}P{1.2cm}P{1.1cm}P{1.1cm}P{1.1cm}}
\begin{tabular}{l|l|l|l}
\toprule[1.2pt]
Dataset & Description &\# split & \# Metrics \\
\midrule
VQAv2~\cite{goyal2017making} & General VQA. General questions about entities, colors, materials, etc. & test-dev & VQA Score \\
\hline
GQA~\cite{hudson2019gqa} & \makecell[l]{Compositional VQA. Built on the scene-graphs in Visual Genome~\cite{krishna2016visual}. \\ More compositional questions and spatial relation questions.} & test-dev & EM  \\
\hline
OK-VQA~\cite{marino2019okvqa} & Knowledge-based VQA. Questions that need external knowledge to be answered. & val & VQA Score   \\
\hline
A-OKVQA~\cite{schwenk2022aokvqa} & An advanced version of OK-VQA that is more challenging. &  &  \\
& \ \ -- Multiple Choice (MC): choose 1 of the 4 options. & val, test & EM \\
& \ \ -- Direct Answer (DA): compare with 10 free-form human answers & val, test & VQA Score  \\
\hline
TallyQA~\cite{acharya2019tallyqa} & Counting questions. & &\\
& \ \ -- Simple: synthesized simple counting questions & test-simple & EM \\
& \ \ -- Complex: human-written complex counting questions & test-complex & EM \\
\hline
TextVQA ~\cite{singh2019towards} & VQA on images that contain text & val & VQA Score \\
\hline
POPE ~\cite{li2023pope} & \makecell[l]{Benchmark on VLM hallucination. \\ Binary questions of whether an object exists in the image.} & dev & EM \\
\hline
MMBench ~\cite{liu2023mmbench} & \makecell[l]{Comprehensive benchmark on VLMs with multiple-choice questions.\\ Covering 20 ability dimensions across 3 levels (e.g., coarse perception, fine-grained \\ perception, attribute reasoning, relation reasoning, logic reasoning, etc.)} & dev & EM\\
\bottomrule[1.2pt]
\end{tabular}
\vspace{-1ex}
\caption{
Summary of evaluation benchmarks.
}
\label{tab:evaluation_data}
% \bottomrule
\vspace{-3mm}
\end{table*}

\subsection{Training details}
We use LoRA~\cite{hu2021lora} to fine-tune both PaLI-3~\cite{chen2023pali3} and PaLI-X~\cite{chen2023palix}. 
For generalist training, we add LoRA weights on each linear layer in the attention blocks and multilayer perceptron (MLP) blocks for both the encoder and decoder in the UL2 transformer. For both models, we use $rank=8$. We use a cosine learning rate schedule, with warm-up ratio $1\%$ and peak learning rate $1e-4$. For all models and all settings, we use a batch size of $128$ and fine-tune the pre-trained model for $8,000$ steps. In terms of training time, we train PaLI-X-VPD with 128 TPU-v3~\cite{Jouppi2017tpu} and it takes about $2$ days to finish training. For \model{PaLI-3-VPD}, we use $32$ TPU-v4 and training takes about $20$ hours. We still observe a steady loss drop when we terminate training, which indicates that more computation may lead to even better performance.
For per-task fine-tuning, to avoid overfitting, we reduce the number of training parameters. For both models, we only add LoRA weights to encoder layers. We use LoRA $rank=4$ for \model{PaLI-X-VPD} and $rank=8$ for \model{PaLI-3-VPD}. The peak learning rate is $1e-4$ and we use a cosine learning schedule, with warm-up ratio $1\%$. For all per-task fine-tuning experiments, we use a batch size of $64$. We train for $1$ epoch on GQA, and $3$ epochs on all other datasets. 
We use the AdamW~\cite{Kingma2015AdamAM} optimizer with $\beta_1 = 0.9$ and $\beta_2 = 0.98$, and bfloat16 for all experiments.

\subsection{Inference costs}
We sampled $300$ questions and measured the computation cost. Using 128 TPU v5, code generation takes on average $4.7s$, while program execution takes $4.2s$. With the same resources, \model{PaLI-X-VPD} takes $0.8s$ per question. \textbf{The cost gap ($0.8s$ vs $8.9s$) is very large} with immense consequences for practical applications.

% \subsection{Program filtering cut}
% As we discussed in \S\ref{sec:intro}, visual programs are error-prone for a variety of reasons. On one hand, the LLM-generated code might contain mistakes—which sometimes lead the program execution to fail (i.e., syntax mistakes), while in other cases contain reasoning mistakes that could pass undetected. On the other hand, even when the high level logic is correct, any of the modules or tools used as subroutines may return erroneous outputs (e.g., the object detection function fails to detect an object). Such errors are not recoverable during execution and are likely to lead to a wrong answer.
% With all these potential sources of error compounded, the overall performance of visual programs is much lower than that of supervised fine-tuned models.

% Specifically, we give PaLM-2~\cite{anil2023palm2} the question $q$, the human-rated ground truth answer $y$, the program execution trace $t_i$, and the program output $z_i$, and let it determine if the trace is reasonable, and the output matches the ground truth (\otilia{example in Appendix X}).

% \subsection{Distill step-by-step cut}
% We refer to ``fine-tuning'' in the broader sense—this can be done via parameter-efficient methods such as soft prompt-tuning~\cite{lester2021power} or LoRA~\cite{hu2021lora}, or end-to-end.

% \subsection{Data mixture for specialist models}
% \label{sec:appendix:training_details:data-mixing-academic}

\section{Human evaluation}
\label{sec:appendix:human_annotation}

We asked our human annotators to first evaluate each model's answer, using the criteria described in \S\ref{sec:experiment:human_eval}. After rating each model answer separately, we also asked them to choose a preferred answer between the two. However, we observed that there are cases where one or both models have similar answers, or both answers are incorrect, in which case it would be difficult for the annotators to choose a favorite, so we also provided the annotators with the options ``Both'' or ``Neither'', giving them the following instruction:  \prompt{Please try to choose "Answer 1 is better" or "Answer 2 is better" whenever possible. We also give you the option to choose "Both are equally good." or "Both are too bad to make a choice." for the cases when it is hard to make a choice either because both answers are correct and similar, or because both answers are wrong so it makes no sense to choose a favorite.}

We show some examples from our human evaluation in Table~\ref{tab:human_annotation_examples}. The table contains the images and corresponding text queries (column 2), the answers provided by the two models we compared—\model{PaLI-X} \model{Instruct} (column 3) and \model{PaLI-X-VPD} (column 4)—along with the corresponding annotations given by the human annotators. The human annotations are aggregated across 3 raters per sample. Finally, column 5 shows which of the two answers was preferred by the human raters. When a model's answer includes a bounding box, we annotate it on the image for convenience.
Examples are as follow:

\begin{itemize}[labelindent=1em]
    \setlength\itemsep{1em}
    \item Example $\#1$ shows a common situation where \model{PaLI-X-VPD} succeeds where \model{PaLI-X} \model{Instruct} fails. By being trained with programs that include calls to an object detection tool, \model{PaLI-X-VPD} has learned to produce answers that localize the object in question in the image, which prods the model to correctly perform tasks such as counting.
    \item Example $\#2$ shows a type of question where neither model produces an explanation, where one is arguably not necessary. However, in spite the lack of explanation, \model{PaLI-X-VPD}'s answer is more accurate.
    \item Example $\#3$ shows an example where from Tally QA where \model{PaLI-X-VPD} is able to correctly localize and count the number of tennis balls in the image, as asked by the question, when \model{PaLI-X} \model{Instruct} counts incorrectly and also does not provide a rationale.
    \item Example $\#4$ shows a situation where \model{PaLI-X-VPD} produces an answer that is deemed both factual and consistent by the human raters, and yet the answer is incorrect. Indeed we can notice that everything that \model{PaLI-X-VPD} is accurate with respect to the context image, and yet it does not address the question as expected by the human raters.
    \item Example $\#5$ shows a case where both models are incorrect, but the annotators prefer \model{PaLI-X-VPD}'s answer, despite the fact that both models essentially give the same incorrect final count of $0$ yellow busses. However, \model{PaLI-X-VPD} provides a consistent rationale, from which we can tell the reason for its mistake: it correctly identifies 2 out of the 3 buses in the image, missing the partially occluded yellow bus in the back. It seems the annotators prefer this kind of mistake over one without an explanation.
\end{itemize}

\begin{table}[H]
\small
\centering
\vspace{-2em}
\begin{tabular}{|c|c|c|c|p{15mm}|}
\hline
    Index &
    Image \& Query & 
    \begin{minipage}[t]{40mm} \centering \model{PaLI-Instruct} Answer \\ \& Human Eval \end{minipage} &
    \begin{minipage}[t]{40mm} \centering \model{PaLI-VPD} Answer \\ \& Human Eval \end{minipage} & 
    Annotator Preference \\
\hline
        1.
    &
        \begin{minipage}[t]{50mm} 
        \raisebox{-\height}{\includegraphics[width=\linewidth]{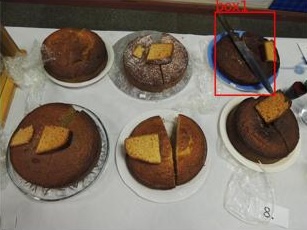}} \\
        On which side is the knife?
        \vspace{1ex}
        \end{minipage}
    & 
        \begin{minipage}[t]{45mm} 
         The knife is on the left side of the table. \\
         \begin{itemize}[noitemsep,nolistsep]
             \item Correct: \xmark
             \item Has Explanation: \xmark
             \item Explanation Factuality: N/A
             \item Explanation Consistency: N/A
         \end{itemize}
        \end{minipage}
    & 
        \begin{minipage}[t]{45mm} 
        The knife is at 55 693 412 884, which is on the right side of the image. \\
        \begin{itemize}[noitemsep,nolistsep]
             \item Correct: \cmark
             \item Has Explanation: \cmark
             \item Explanation Factuality: \cmark
             \item Explanation Consistency: \cmark
         \end{itemize}
        
        \end{minipage}
    &
        \model{PaLI-VPD}
    \\
\hline
        2.
    &
        \begin{minipage}[t]{50mm} 
        \raisebox{-\height}{\includegraphics[width=\linewidth]{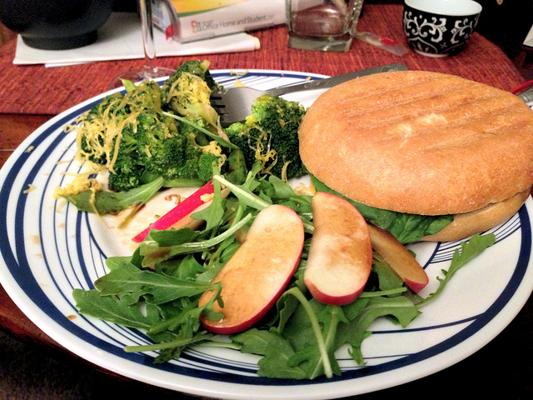}} \\
        What are this fruits called?
        \vspace{1ex}
        \end{minipage}
    & 
        \begin{minipage}[t]{45mm} 
         The fruits in the image are apples and broccoli. \\
         \begin{itemize}[noitemsep,nolistsep]
             \item Correct: \xmark
             \item Has Explanation: \xmark
             \item Explanation Factuality: N/A
             \item Explanation Consistency: N/A
         \end{itemize}
        \end{minipage}
    & 
        \begin{minipage}[t]{45mm} 
        The fruits are called apples. \\
        \begin{itemize}[noitemsep,nolistsep]
             \item Correct: \cmark
             \item Has Explanation: \xmark
             \item Explanation Factuality: N/A
             \item Explanation Consistency: N/A
         \end{itemize}
        
        \end{minipage}
    &
        \model{PaLI-VPD}
    \\
\hline
        3.
    &
        \begin{minipage}[t]{45mm} 
        \raisebox{-\height}{\includegraphics[width=0.6\linewidth]{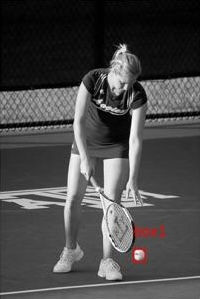}} \\
        How many tennis balls are in the photo?
        \vspace{1ex}
        \end{minipage}
    & 
        \begin{minipage}[t]{45mm} 
        There are two tennis balls in the photo.\\
         \begin{itemize}[noitemsep,nolistsep]
             \item Correct: \xmark
             \item Has Explanation: \xmark
             \item Explanation Factuality: N/A
             \item Explanation Consistency: N/A
         \end{itemize}
        \end{minipage}
    & 
        \begin{minipage}[t]{45mm} 
        There is a tennis ball at 826 665 869 721. Thus, there is 1 tennis ball.  \\
        \begin{itemize}[noitemsep,nolistsep]
             \item Correct: \cmark
             \item Has Explanation: \cmark
             \item Explanation Factuality: \cmark
             \item Explanation Consistency: \cmark
         \end{itemize}
        
        \end{minipage}
    & 
        \begin{minipage}[t]{10mm} \model{PaLI-VPD} \end{minipage}
    \\
\hline
        4.
    &
        \begin{minipage}[t]{50mm} 
        \raisebox{-\height}{\includegraphics[width=\linewidth]{figures/examples/aokvqa_direct_YgiqkypFRQHVYHqieDPN9u.jpeg}} \\
        Why is one player wearing gloves?
        \vspace{1ex}
        \end{minipage}
    & 
        \begin{minipage}[t]{45mm} 
         The player is wearing gloves because he is a batter. Batters wear gloves to protect their hands from the ball.	 \\
         \begin{itemize}[noitemsep,nolistsep]
             \item Correct: \cmark
             \item Has Explanation: \xmark
             \item Explanation Factuality: \cmark
             \item Explanation Consistency: \cmark
         \end{itemize}
        \end{minipage}
    & 
        \begin{minipage}[t]{45mm} 
        The player is wearing gloves to catch the ball.	 \\
        \begin{itemize}[noitemsep,nolistsep]
             \item Correct: \xmark
             \item Has Explanation: \cmark
             \item Explanation Factuality: \cmark
             \item Explanation Consistency: \cmark
         \end{itemize}
        
        \end{minipage}
    & 
        \begin{minipage}[t]{10mm} \model{PaLI-\\Instruct} \end{minipage}
    \\
    
\hline
        5.
    &
        \begin{minipage}[t]{50mm} 
        \raisebox{-\height}{\includegraphics[width=\linewidth]{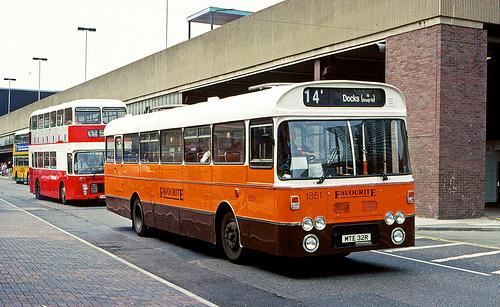}} \\
        How many yellow buses are in the picture?
        \vspace{1ex}
        \end{minipage}
    & 
        \begin{minipage}[t]{45mm} 
        There are no yellow buses in the picture.	 \\
         \begin{itemize}[noitemsep,nolistsep]
             \item Correct: \xmark
             \item Has Explanation: \xmark
             \item Explanation Factuality: N/A
             \item Explanation Consistency: N/A
         \end{itemize}
        \end{minipage}
    & 
        \begin{minipage}[t]{45mm} 
        There are 2 buses in the picture. Neither of them is yellow. Thus, there are 0 yellow buses in the picture. \\
        \begin{itemize}[noitemsep,nolistsep]
             \item Correct: \xmark
             \item Has Explanation: \cmark
             \item Explanation Factuality: \xmark
             \item Explanation Consistency: \cmark
         \end{itemize}
        
        \end{minipage}
    & 
        \begin{minipage}[t]{10mm} \model{PaLI-VPD} \end{minipage}
    \\
\hline
\end{tabular}
\vspace{-2ex}
\caption{Human Annotation Examples. We show the answers provided by the two models, \model{PaLI-X} \model{Instruct} and \model{PaLI-X-VPD}, along with the aggregated evaluation scores by human raters along the criteria introduced in \S\ref{sec:experiment:human_eval}.}
\label{tab:human_annotation_examples}
\vspace{-3mm}
\end{table}

\newpage

\newpage

\section{Prompts}
\label{sec:appendix:prompts}
In this section we present the prompts used in our data synthesis pipeline. Refer to \S\ref{sec:appendix:examples} for step-by-step examples of the programs, execution traces, and the converted CoTs.

\subsection{Prompt for code generation}
For each image and query, we put the query and a model-generated image caption in the prompt. An LLM takes this prompt and generate the program to answer the query. We modify the original ViperGPT~\cite{surismenon2023vipergpt} prompt to adapt to the vision tools we use in this paper.

\begin{lstlisting}[language=Python]
class ImagePatch:
    # A Python class containing a crop of an image centered around a particular object, as well as relevant information.
    # Attributes
    # ----------
    # cropped_image : array_like
    #     An array-like of the cropped image taken from the original image.
    # left, lower, right, upper : int
    #     An int describing the position of the (left/lower/right/upper) border of the crop's bounding box in the original image.
    # Methods
    # -------
    # find(object_name: str)->List[ImagePatch]
    #     Returns a list of new ImagePatch objects containing crops of the image centered around any objects found in the
    #     image matching the object_name.
    # visual_question_answering(question: str=None)->str
    #     Returns the answer to a basic question asked about the image. If no question is provided, returns the answer to "What is this?".
    # image_caption()->str
    #     Returns a short description of the image crop.
    # expand_patch_with_surrounding()->ImagePatch
    #     Returns a new ImagePatch object that contains the current ImagePatch and its surroundings.
    # overlaps(patch: ImagePatch)->Bool
    #     Returns True if the current ImagePatch overlaps with another patch and False otherwise
    # compute_depth()->float
    #     Returns the median depth of the image patch. The bigger the depth, the further the patch is from the camera.

    def __init__(self, image, left: int = None, lower: int = None, right: int = None, upper: int = None):
        # Initializes an ImagePatch object by cropping the image at the given coordinates and stores the coordinates as
        # attributes. If no coordinates are provided, the image is left unmodified, and the coordinates are set to the
        # dimensions of the image.
        # Parameters
        # -------
        # image: PIL.Image
        #     An array-like of the original image.
        # left, lower, right, upper : int
        #     An int describing the position of the (left/lower/right/upper) border of the crop's bounding box in the original image.
        #     The coordinates (y1,x1,y2,x2) are with respect to the upper left corner the original image.
        #     To be closer with human perception, left, lower, right, upper are with respect to the lower left corner of the squared image.
        #     Use left, lower, right, upper for downstream tasks.

        self.original_image = image
        size_x, size_y = image.size

        if left is None and right is None and upper is None and lower is None:
            self.x1 = 0
            self.y1 = 0
            self.x2 = 999
            self.y2 = 999
        else:
            self.x1 = left
            self.y1 = 999 - upper
            self.x2 = right
            self.y2 = 999 - lower

        self.cropped_image = image.crop((int(self.x1/1000*self.sz), int(self.y1/1000*self.sz),
                                        int(self.x2/1000*self.sz), int(self.y2/1000*self.sz)))

        self.width = self.x2 - self.x1
        self.height = self.y2 - self.y1

        # all coordinates use the upper left corner as the origin (0,0).
        # However, human perception uses the lower left corner as the origin.
        # So, need to revert upper/lower for language model
        self.left = self.x1
        self.right = self.x2
        self.upper = 999 - self.y1
        self.lower = 999 - self.y2

        self.horizontal_center = (self.left + self.right) / 2
        self.vertical_center = (self.lower + self.upper) / 2

        self.patch_description_string = f"{self.y1} {self.x1} {self.y2} {self.x2}"

    def __str__(self):
        return self.patch_description_string

    def compute_depth(self):
        # compute the depth map on the full image. Returns a np.array with size 192*192
        # Parameters
        # ----------
        # Returns
        # -------
        # float
        #     the median depth of the image crop

        # Examples
        # --------
        # >>> return the image patch of the bar furthest away
        # >>> def execute_command(image)->ImagePatch:
        # >>>     image_patch = ImagePatch(image)
        # >>>     bar_patches = image_patch.find("bar")
        # >>>     bar_patches.sort(key=lambda bar: bar.compute_depth())
        # >>>     return bar_patches[-1]

        return depth(self.cropped_image)
    
    def find(self, object_name: str):
        # Returns a list of ImagePatch objects matching object_name contained in the crop if any are found.
        # The object_name should be as simple as example, including only nouns
        # Otherwise, returns an empty list.
        # Note that the returned patches are not ordered
        # Parameters
        # ----------
        # object_name : str
        #     the name of the object to be found

        # Returns
        # -------
        # List[ImagePatch]
        #     a list of ImagePatch objects matching object_name contained in the crop

        # Examples
        # --------
        # >>> # find all the kids in the images
        # >>> def execute_command(image) -> List[ImagePatch]:
        # >>>     image_patch = ImagePatch(image)
        # >>>     kid_patches = image_patch.find("kid")
        # >>>     return kid_patches

        print(f"Calling find function. Detect {object_name}.")
        det_patches = detect(self.cropped_image, object_name)
        print(f"Detection result: {' and '.join([str(d) + ' ' + object_name for d in det_patches])}")

        return det_patches


    def expand_patch_with_surrounding(self):
        # Expand the image patch to include the surroundings. Now done by keeping the center of the patch
        # and returns a patch with double width and height

        # Examples
        # -------
        # >>> # How many kids are not sitting under an umbrella?
        # >>> def execute_command(image):
        # >>>     image_patch = ImagePatch(image)
        # >>>     kid_patches = image_patch.find("kid")

        # >>>     # Find the kids that are under the umbrella.
        # >>>     kids_not_under_umbrella = []

        # >>>     for kid_patch in kid_patches:
        # >>>         kid_with_surrounding = kid_patch.expand_patch_with_surrounding()
        # >>>         if "yes" in kid_with_surrounding.visual_question_answering("Is the kid under the umbrella?"):
        # >>>             print(f"the kid at {kid_patch} is sitting under an umbrella.")
        # >>>         else:
        # >>>             print(f"the kid at {kid_patch} is not sitting under an umbrella.")
        # >>>             kids_not_under_umbrella.append(kid_patch)

        # >>>     # Count the number of kids under the umbrella.
        # >>>     num_kids_not_under_umbrella = len(kids_not_under_umbrella)

        # >>>     return formatting_answer(str(num_kids_not_under_umbrella))

        new_left = max(self.left - self.width / 2, 0)
        new_right = min(self.right + self.width / 2, 999)
        new_lower = max(self.lower - self.height / 2,0)
        new_upper = min(self.upper + self.height / 2, 999)

        return ImagePatch(self.original_image, new_left, new_lower, new_right, new_upper)


    def visual_question_answering(self, question: str = None) -> str:
        # Returns the answer to a basic question asked about the image.
        # The questions are about basic perception, and are not meant to be used for complex reasoning
        # or external knowledge.

        # Parameters
        # -------
        # question : str
        #     A string describing the question to be asked.

        # Examples
        # -------

        # >>> # What is the name of the player in this picture?
        # >>> def execute_command(image) -> str:
        # >>>     image_patch = ImagePatch(image)
        # >>>     return formatting_answer(image_patch.visual_question_answering("What is the name of the player?"))

        # >>> # What color is the foo?
        # >>> def execute_command(image) -> str:
        # >>>     image_patch = ImagePatch(image)
        # >>>     return formatting_answer(foo_patch.visual_question_answering("What color is the foo?"))

        # >>> # What country serves this kind of food the most?
        # >>> def execute_command(image) -> str:
        # >>>     image_patch = ImagePatch(image)
        # >>>     food_name = image_patch.visual_question_answering("What kind of food is served?")
        # >>>     country = language_question_answering(f"What country serves {food_name} most?", long_answer=False)
        # >>>     return formatting_answer(country)

        # >>> # Is the second bar from the left quuxy?
        # >>> def execute_command(image) -> str:
        # >>>     image_patch = ImagePatch(image)
        # >>>     bar_patches = image_patch.find("bar")
        # >>>     bar_patches.sort(key=lambda x: x.horizontal_center)
        # >>>     bar_patch = bar_patches[1]
        # >>>     return formatting_answer(bar_patch.visual_question_answering("Is the bar quuxy?"))

        answer = vqa(self.cropped_image, question)

        print(f"Calling visual_question_answering function.")
        print(f"Question: {question}")
        print(f"Answer: {answer}")

        return answer

    def image_caption(self) -> str:
        # Returns a short description of the image.
        return image_caption(self.cropped_image)

    def overlaps(self, patch) -> bool:
        # check if another image patch overlaps with this image patch
        # if patch overlaps with current patch, return True. Otherwise return False

        if patch.right < self.left or self.right < patch.left:
          return False
        if patch.lower > self.upper or self.lower > patch.upper:
          return False
        return True


def language_question_answering(question: str, long_answer: bool = False) -> str:
    # Answers a text question using a lanugage model like PaLM and GPT-3. The input question is always a formatted string with a variable in it.
    # Default is short-form answers, can be made long-form responses with the long_answer flag.

    # Parameters
    # ----------
    # question: str
    #     the text question to ask. Language model cannot anderstand the image. Must not contain any reference to 'the image' or 'the photo', etc.
    # long_answer: bool
    #     whether to return a short answer or a long answer. Short answers are one or at most two words, very concise.
    #     Long answers are longer, and may be paragraphs and explanations. Defalt is False.

    # Examples
    # --------
    # >>> # What is the city this building is in?
    # >>> def execute_command(image) -> str:
    # >>>     image_patch = ImagePatch(image)
    # >>>     building_name = building_patch.visual_question_answering("What is the name of the building?")
    # >>>     return formatting_answer(language_question_answering(f"What city is {building_name} in?", long_answer=False))

    # >>> # Who invented this object?
    # >>> def execute_command(image) -> str:
    # >>>     image_patch = ImagePatch(image)
    # >>>     object_name = object_patch.visual_question_answering("What is this object?")
    # >>>     return formatting_answer(language_question_answering(f"Who invented {object_name}?", long_answer=False))

    # >>> # Explain the history behind this object.
    # >>> def execute_command(image) -> str:
    # >>>     image_patch = ImagePatch(image)
    # >>>     object_name = object_patch.visual_question_answering("What is the object?")
    # >>>     return formatting_answer(.language_question_answering(f"What is the history behind {object_name}?", long_answer=True))
    print(f"Calling language_question_answering")
    print(f"Question: {question}")

    answer = language_model_qa(question, long_answer).lower().strip()
    print(f"Answer: {answer}")
    return answer


def distance(patch_a: Union[ImagePatch, float], patch_b: Union[ImagePatch, float]) -> float:
    # Returns the distance between the edges of two ImagePatches, or between two floats.
    # If the patches overlap, it returns a negative distance corresponding to the negative intersection over union.

    # Parameters
    # ----------
    # patch_a : ImagePatch
    # patch_b : ImagePatch

    # Examples
    # --------
    # # Return the qux that is closest to the foo
    # >>> def execute_command(image):
    # >>>     image_patch = ImagePatch(image)
    # >>>     qux_patches = image_patch.find('qux')
    # >>>     foo_patches = image_patch.find('foo')
    # >>>     foo_patch = foo_patches[0]
    # >>>     qux_patches.sort(key=lambda x: distance(x, foo_patch))
    # >>>     return qux_patches[0]

    return dist(patch_a, patch_b)


def formatting_answer(answer) -> str:
    # Formatting the answer into a string that follows the task's requirement
    # For example, it changes bool value to "yes" or "no", and clean up long answer into short ones.
    # This function should be used at the end of each program

    final_answer = ""
    if isinstance(answer, str):
      final_answer = answer.strip()

    elif isinstance(answer, bool):
      final_answer = "yes" if answer else "no"

    elif isinstance(answer, list):
      final_answer = ", ".join([str(x) for x in answer])

    elif isinstance(answer, ImagePatch):
      final_answer = answer.image_caption()

    else:
      final_answer = str(answer)

    print(f"Program output: {final_answer}")
    return final_answer


Given an image and a query, write the function execute_command using Python and the ImagePatch class (above), and the other functions above that could be executed to provide an answer to the query.
For reference, a model generated image description is also provided, so that the function can be customized for the given image. The image description is model-generated and may not be reliable, so do not trust it.

Consider the following guidelines:
- Use base Python (comparison, sorting) for basic logical operations, left/right/up/down, math, etc.
- Use the language_question_answering function to access external information and answer informational questions NOT concerning the image.
- The program should print out the intermediate traces as it runs. So add print function in the program if needed.

For usual cases, follow the guidelines below:
- For simple visual queries, directly call visual_question_answering to get the answer.
- For queries that need world knowledge, commonsense knowledge, and language reasoning, use visual_question_answering, language_question_answering, and sometimes image_caption to get the answer.
- For queries that require counting and spatical relations, in addition to the above functions, use find function to help getting the answer.
- For queries involve "behind" and "front", consider using compute_depth function.


Some examples:
Image description: a woman is walking several dogs
Query: how many dogs are to left of the person?
Function:
def execute_command(image):
    image_patch = ImagePatch(image)
    person_patch = image_patch.find("person")[0]
    dog_patches = image_patch.find("dog")

    # Count the number of dogs whose leftmost x-coordinate is less than the person.
    num_dogs_left = 0
    for dog_patch in dog_patches:
        if dog_patch.left < person_patch.horizontal_center:
            print(f"dog at {dog_patch} is on the left of human.")
            num_dogs_left += 1

    return formatting_answer(num_dogs_left)

# [other in-context examples]

Image description: INSERT_IMAGE_CAPTION_HERE
Query: INSERT_QUERY_HERE
Function:
\end{lstlisting}

\subsection{Prompt for result verification}
After running each program, we get an output. As discussed in \S\ref{sec:method:generate_cot}, we adopt the method of~\cite{kamalloo2023evaluating} and use an LLM to determine if the program output matches human answers. 
The LLM takes the program output and reference answers as input. The prompt is as follows:

\begin{lstlisting}
Given a visual question, several human annotator answers, and a candidate answer, determine if the candidate is correct.
The candidate is considered correct if  is allowed to have formatting differences compared with the human answers.
If the candidate is correct, return the gold answer it matches. Otherwise, return None.

Question: INSERT_QUESTION_HERE
Answers: INSERT_ANSWERS_HERE
Candidate: INSERT_CANDIDATE_HERE
Is the candidate correct?
\end{lstlisting}

\subsection{Prompt for CoT conversion}
Finally, once a program is filtered, we convert its execution trace into chain-of-thought using an LLM.
The LLM takes the query, program, execution trace, program output as input, and summarizes the execution trace into a chain-of-thought rationale.
The prompt we use as as follows:

\begin{lstlisting}
Given an image and a question, I wrote the function execute_command using Python and the ImagePatch class (above), and the other functions above that could be executed to provide an answer to the query.
As shown in the code, the code will print execution traces.
I need you to rewrite the execution trace into a natural language rationale that leads to the answer.

Consider the following guidelines:
- Use the bounding box information in the rationale.
- Referencing the execution trace, write a reasoning chain that leads to the most common human answer. Notice that the output should be the same as the human answer, not necessarily the program output.


Some examples:
Question: How many wheels does the plane have?
Program:
def execute_command(image):
    image_patch = ImagePatch(image)

    # Find the plane in the image
    plane_patch = image_patch.find("plane")[0]

    # Count the number of wheels on the plane
    num_wheels = 0
    for wheel in plane_patch.find("wheel"):
        num_wheels += 1

    return formatting_answer(str(num_wheels))
Execution trace:
Calling find function. Detect plane
Detected plane at 153 25 647 972
Calling find function. Detect wheel
Detected wheel at 603 471 649 515
Detected wheel at 621 85 646 113
Detected wheel at 615 383 645 428
Program output: 3
Rationale: The plane at 153 25 647 972 has wheels at 603 471 649 515, 621 85 646 113, and 615 383 645 428.Thus, it has 3 wheels.

[Other demonstration examples]

Question: INSERT_QUESTION_HERE
INSERT_PROGRAM_HERE
Execution trace:
INSERT_EXECUTION_TRACE_HERE
Rationale:
\end{lstlisting}

\end{document}